%% file: ms.tex
\documentclass[10pt,twocolumn,letterpaper]{article}

\usepackage{iccv}
\usepackage{times}
\usepackage{epsfig}
\usepackage{graphicx}
\usepackage{amsmath}
\usepackage{amssymb}
\usepackage{multirow}
\usepackage[flushleft]{threeparttable}
\usepackage{subfig}
\usepackage{placeins}
\usepackage{float} 
\usepackage{enumitem}
\usepackage{color}
\usepackage{svg}
\usepackage{cleveref}
\usepackage[pagebackref=true,breaklinks=true,letterpaper=true,colorlinks,bookmarks=false]{hyperref}
\usepackage{booktabs}
\usepackage{balance}

\usepackage{sistyle}
\SIthousandsep{,}

\usepackage{xparse}
\ExplSyntaxOn
\DeclareExpandableDocumentCommand{\eval}{m}{\int_eval:n {#1}}
\DeclareExpandableDocumentCommand{\feval}{m}{\fp_eval:n {#1}}
\ExplSyntaxOff

\newcommand{\VQATrain}{\eval{9772}}
\newcommand{\VQAVal}{\eval{1504}}
\newcommand{\VQATest}{\eval{3758}}
\newcommand{\VQATotal}{\eval{\VQATrain+\VQAVal+\VQATest}}

\newcommand{\VizWizTrain}{\eval{19196}}
\newcommand{\VizWizVal}{\eval{3048}}
\newcommand{\VizWizTest}{\eval{7677}}
\newcommand{\VizWizTotal}{\eval{\VizWizTrain+\VizWizVal+\VizWizTest}}

\newcommand{\VizWizMLTrain}{\eval{19969}}
\newcommand{\VizWizMLVal}{\eval{3166}}
\newcommand{\VizWizMLTest}{\eval{7983}}
\newcommand{\VizWizMLTotal}{\eval{\VizWizMLTrain+\VizWizMLVal+\VizWizMLTest}}

\newcommand{\VQsCrowdsourced}{\eval{\VizWizTotal+\VQATotal}}
\newcommand{\AnnsCrowdsourced}{\eval{5*\eval{\VQsCrowdsourced}}}

\iccvfinalcopy 

\def\iccvPaperID{1111} 

\ificcvfinal\fi

\begin{document}

\title{Why Does a Visual Question Have Different Answers?}

\author{Nilavra Bhattacharya$^*$, Qing Li$^+$, Danna Gurari$^*$ \\
\noindent
{\small $~^*$ University of Texas at Austin},
{\small $~^+$ University of California, Los Angeles} 
}

\maketitle

\begin{abstract} 
Visual question answering is the task of returning the answer to a question about an image.  A challenge is that different people often provide different answers to the same visual question.  To our knowledge, this is the first work that aims to understand why.  We propose a taxonomy of nine plausible reasons, and create two labelled datasets consisting of $\sim$45,000 visual questions indicating which reasons led to answer differences.  We then propose a novel problem of predicting directly from a visual question which reasons will cause answer differences as well as a novel algorithm for this purpose.  Experiments demonstrate the advantage of our approach over several related baselines on two diverse datasets.  We publicly share the datasets and code at \texttt{https://vizwiz.org}.
\end{abstract}

\input{introduction}
\input{related-works}
\input{dataset-creation}
\input{dataset-analysis}
\input{prediction-system}
\input{conclusions}

\vspace{0.5em}
\noindent
\textbf{Acknowledgements.}
\noindent
We thank the anonymous reviewers for their valuable feedback and the crowd workers for providing the annotations.  This work is supported in part by National Science Foundation funding (IIS-1755593).

\balance{}
{\small
\bibliographystyle{ieee_fullname}
\bibliography{ms}
}

\clearpage
\input{supp-materials.tex}

\end{document}

%% file: introduction.tex
\section{Introduction}
\label{sec:intro}
Visual question answering (VQA), the task of returning the answer to a question about an image, is of widespread interest across both industry and academia.  For example, many people who are blind find existing VQA solutions are indispensable assistants for providing answers to their daily visual questions; e.g., they use vision-based assistants such as VizWiz~\cite{bigham2010VizWiznearlyrealtime} and BeSpecular~\cite{BeSpecular} to snap a photo with their mobile phones and then receive answers from remote workers.  In the artificial intelligence research community, the VQA problem has emerged as an iconic challenge for emulating a human's vision and language capabilities~\cite{antol2015VqaVisualquestion,kafle2017analysisvisualquestion}.

Despite the tremendous social impact and progress with VQA solutions, a limitation is that most solutions lack a way to handle when a visual question elicits different answers from different people.  The prevailing assumption is that the goal is to \emph{return a single answer}.  Yet prior work~\cite{gurari2017CrowdVergePredictingIf} has shown that it is common for visual questions to elicit different answers---it occurs for over half of nearly 500,000 visual questions in three VQA datasets.  

\begin{figure*} [thbp]
\fbox{\includegraphics[clip=true, trim=0 10 45 5, width=\textwidth]{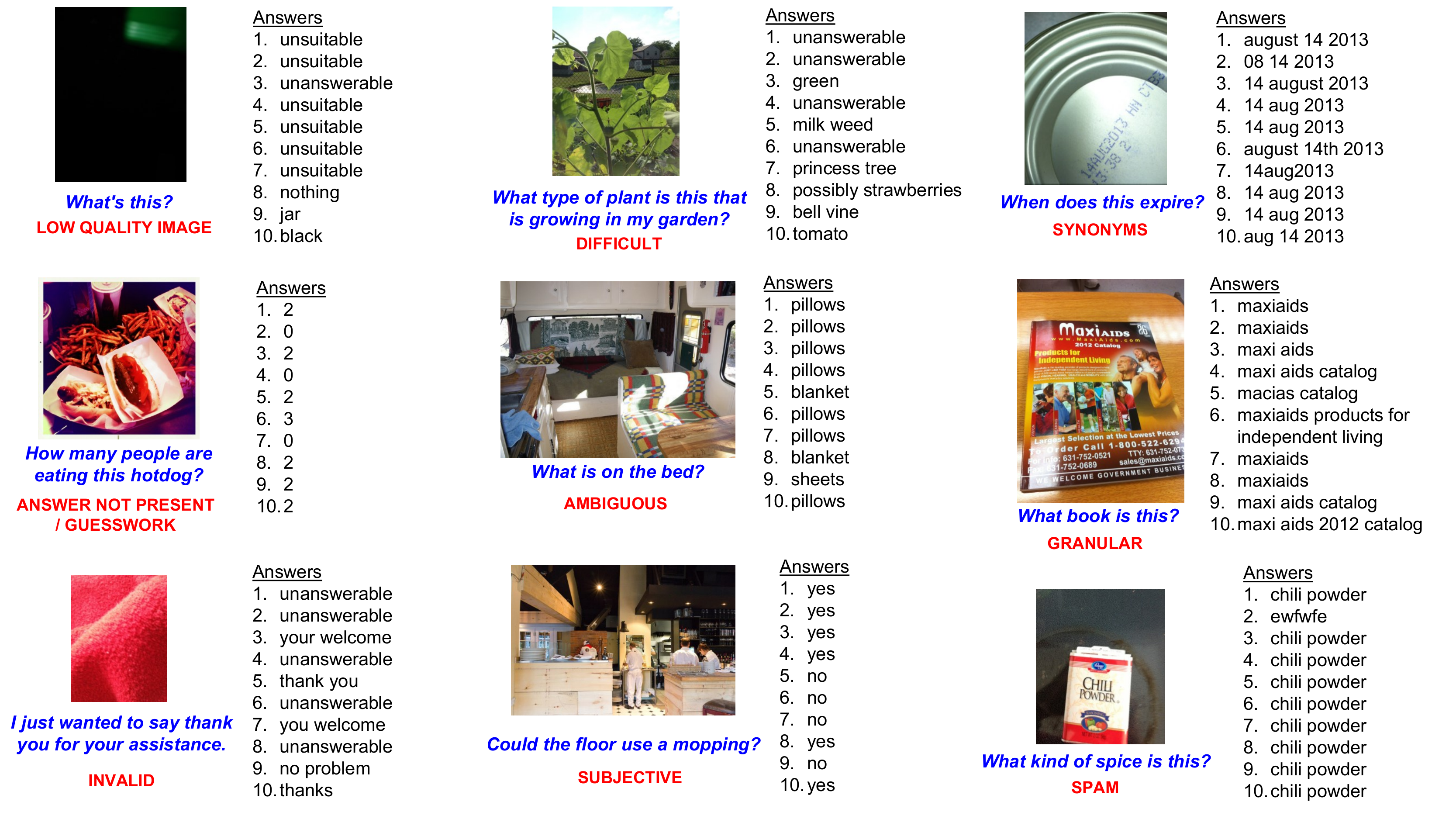}} 
\vspace{-1.2em}
\caption{Examples of visual questions (VQs) asked by people who are blind and sighted, and answers from 10 different people.  As shown, the answers can differ for a variety of reasons, including because of the VQ (first and second columns) or the answers (third column).  We propose a novel problem of predicting why answers will differ for a VQ and a solution.}
  \label{fig:disagreement_examples}
\end{figure*}

Our goal is to identify why different answers arise.  Accordingly, our work is premised on the assumption that there can \emph{exist multiple answers} for a visual question.  Our work extends prior work~\cite{antol2015VqaVisualquestion,gurari2017CrowdVergePredictingIf,malinowski2015Askyourneurons,teney2018Tipstricksvisual} which has suggested reasons why answers can differ---including that visual questions are difficult, subjective, or ambiguous as well as synonymous answers.  First, we propose a taxonomy of nine plausible reasons why answers can differ, which are exemplified in Figure~\ref{fig:disagreement_examples}.  We next ask crowd workers to identify which of these reasons led to answer differences for each of $\sim$45,000 visual questions asked by people who are blind and sighted.  Finally, we propose a novel problem of predicting which reasons will lead to answer differences directly from a visual question and propose a novel algorithm for this task.  Our findings offer compelling evidence that an algorithmic framework must learn to simultaneously model and synthesize different individuals' (potentially conflicting) perceptions of images and language.  

We offer our work as a valuable foundation for improving VQA services, by empowering system designers and users to know how to prevent, interpret, or resolve answer differences.  Specifically, a solution that anticipates why a visual question will lead to different answers (summarized in Figure~\ref{fig:disagreement_examples}) could (1) help users identify how to modify their visual question in order to arrive at a single, unambiguous answer; e.g., retake an image when it is low quality or does not show the answer versus modify the question when it is ambiguous or invalid; (2) increase users' awareness for what reasons, if any, trigger answer differences when they are given a single answer; or (3) reveal how to automatically aggregate different answers~\cite{amid2015Multiviewtripletembedding,gurari2018PredictingForegroundObject,jas2015Imagespecificity,kovashka2015Discoveringattributeshades,wan2016ModelingAmbiguitySubjectivity} when multiple answers are collected.

More generally, to our knowledge, this is the first work in the computer vision community to characterize, quantify, and model reasons why annotations differ.  We believe it will motivate and facilitate future work on related problems that similarly face annotator differences, including image captioning, visual storytelling, and visual dialog.  We publicly share the datasets and code to encourage community progress in developing algorithmic frameworks that can account for the diversity of perspectives in a crowd.

%% file: related-works.tex
\section{Related Work}
\label{sec:bg}
\paragraph{Visual Question Answering Datasets.}
Numerous dataset challenges have been posed to encourage the development of algorithms that automatically answer visual questions~\cite{antol2015VqaVisualquestion,goyal2017MakingVQAmatter,gurari2018VizWizGrandChallenge,kafle2017analysisvisualquestion}.  The shared goal of these challenges is to return a single answer to each visual question.  Yet, visual questions often lead to different answers from different people~\cite{antol2015VqaVisualquestion,gurari2017CrowdVergePredictingIf,malinowski2015Askyourneurons,teney2018Tipstricksvisual}.  Prior work~\cite{antol2015VqaVisualquestion,malinowski2015Askyourneurons} tried to mitigate this problem on the performance metric side, by using consensus metrics.  We instead introduce the first VQA dataset that fosters research in learning why different answers will arise.  Specifically, for two popular VQA datasets, VizWiz~\cite{gurari2018VizWizGrandChallenge} and VQA\_2.0~\cite{antol2015VqaVisualquestion}, we label each visual question with metadata indicating which among nine options are the reasons for the observed answer differences.  Experiments demonstrate these datasets are valuable for training algorithms to predict why answers will differ for any visual question.  

\vspace{-1.25em}
\paragraph{Challenges/Obstacles for Answering Visual Questions.}
Our work relates to the body of literature aimed at understanding what can make a visual question challenging, or even impossible, to answer.  One work examined the issue of \emph{difficulty} (aka - required skill level), by identifying the minimum age needed to successfully answer a visual question~\cite{antol2015VqaVisualquestion}.  Another set of works explored the issue of \emph{relevance}, and in particular identifying when questions are unrelated to the contents of images~\cite{mahendru2017promise}.  Another work examined the issue of \emph{answerability}, with an emphasis on when questions cannot be answered due to extreme image quality issues including blur, saturation, and fingers obstructing the camera view~\cite{gurari2018VizWizGrandChallenge}.  Our work complements prior work in that we found that each of these issues are commonly associated with visual questions that evoke different answers; e.g., see examples in Figure~\ref{fig:disagreement_examples} for ``Difficult", ``Answer Not Present", and ``Low Quality Image" respectively.  Our experiments demonstrate a strong advantage of designing algorithms to directly predict the reason why answers will differ over relying on related methods that predict \emph{relevance}~\cite{mahendru2017promise} or \emph{answerability}~\cite{gurari2018VizWizGrandChallenge} alone.

\begin{table*}[t!]
\centering
\includegraphics[clip=true, trim=0 62 65 0, width=0.96\linewidth]{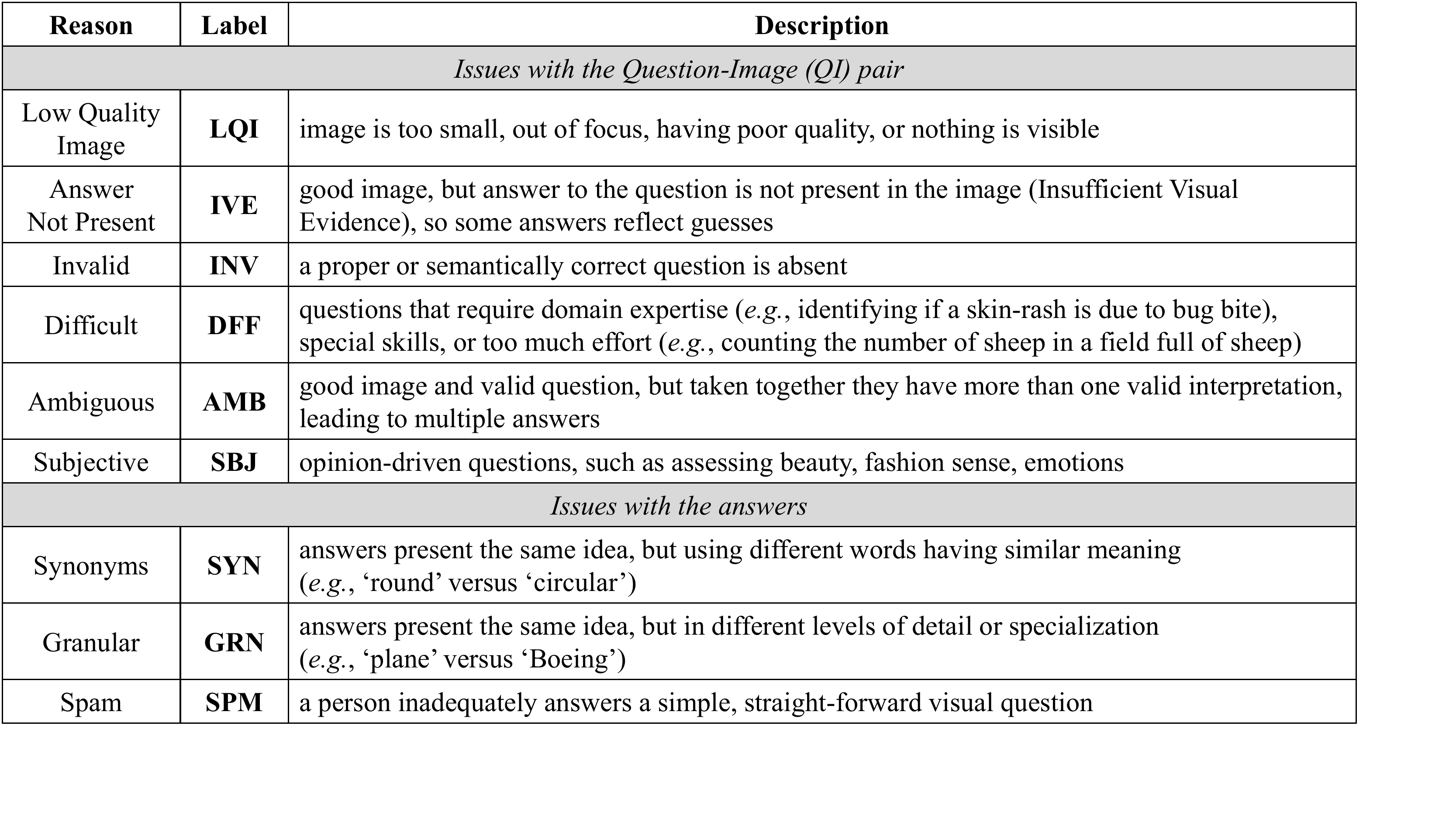} 
\vspace{-0.5em}
\caption{Proposed taxonomy of nine reasons that evoke differing answers for visual questions.}
\label{tab:tab_taxonomy}
\end{table*}

\vspace{-1.25em}
\paragraph{Understanding Why Crowd Responses Differ.}
More broadly, related work includes literature about why a crowd's annotations can differ and how to resolve those differences (largely in domains outside VQA).  Commonly, it is assumed there is a single true answer and that any observed differences stem from insufficient annotator performance, for example because the task is too difficult for some annotators~\cite{amirkhani2014Agreementdisagreementbased, aroyo2013CrowdTruthHarnessing, dumitrache2017Crowdsourcinggroundtruth,inel2013Domainindependentqualitymeasures,inel2014CrowdtruthMachinehumancomputation,sharmanska2016AmbiguityHelpsClassification,sheshadri2013Squarebenchmarkresearch,soberon2013Measuringcrowdtruth,welinder2010multidimensionalwisdomcrowds} or
because malicious workers submit spam~\cite{eickhoff2013Increasingcheatrobustness, gadiraju2015Understandingmaliciousbehavior,vuurens2012obtaining}.  Others have embraced the possibility that the task may be ambiguous~\cite{amid2015Multiviewtripletembedding,gurari2018PredictingForegroundObject,jas2015Imagespecificity,kovashka2015Discoveringattributeshades} or subjective~\cite{wan2016ModelingAmbiguitySubjectivity} and so multiple annotations can be valid.  While each work embeds assumptions regarding why answers differ, a challenge remains of knowing which assumptions are valid when.  Thus, we conduct a systematic study to enumerate plausible reasons and propose an algorithm to decide which reasons apply when.  

\vspace{-1.25em}
\paragraph{Learning to Anticipate Annotation Differences.}
Related works have trained algorithms to anticipate when annotation differences will arise for numerous vision problems.  Some methods recognize when visual content is \emph{ambiguous} and so will lead to diverse human interpretations, including for captioning images~\cite{jas2015Imagespecificity}, interpreting visual attributes~\cite{kovashka2015Discoveringattributeshades}, and locating the most prominent foreground object~\cite{gurari2018PredictingForegroundObject}.  Other methods recognize to what extent visual content will evoke people's \emph{subjective} perceptions, including for visual humor~\cite{chandrasekaran2016Wearehumor} and memorability~\cite{isola2011Whatmakesimage}.  Other methods anticipate whether a crowd will offer different responses to a visual question~\cite{gurari2017CrowdVergePredictingIf} and to what extent~\cite{yang2018VisualQuestionAnswer}.  Complementing prior work, we propose the first solution for deciphering which of multiple reason(s) will lead to annotation differences.  A key insight for our approach is to employ a VQA system's predicted answers as predictive cues, motivated by the belief that VQA models trained to optimize for multiple correct answers (e.g., using ``soft target" scores) embed some understanding for why there may be uncertainty around a single true answer~\cite{teney2018Tipstricksvisual}.  

%% file: dataset-creation.tex
\section{Labeled Datasets}
\label{sec:dataset_creation}
We now introduce how we created the datasets, which consist of VQAs paired with labels indicating why answers differ.  In particular, in this section, we describe our taxonomy and data labeling approach.

\vspace{-0.75em}
\paragraph{Reasons for Differing Answers.}
We developed a taxonomy of nine reasons for why answers may differ, which are summarized in Table~\ref{tab:tab_taxonomy}.
Six of the nine reasons are inspired by the crowdsourcing literature -- INV~\cite{michael2017crowdsourcing}, DFF~\cite{welinder2010multidimensionalwisdomcrowds}, AMB~\cite{jas2015Imagespecificity,kovashka2015Discoveringattributeshades, wan2016ModelingAmbiguitySubjectivity}, SBJ~\cite{nguyen2016Probabilisticmodelingcrowdsourcing, wan2016ModelingAmbiguitySubjectivity, burton2012crowdsourcing}, 
SYN~\cite{michael2017crowdsourcing}, and SPM~\cite{vuurens2011much, vuurens2012obtaining, eickhoff2013Increasingcheatrobustness, gadiraju2015Understandingmaliciousbehavior}.  Two of the reasons are inspired by prior visual question answering work~\cite{gurari2018VizWizGrandChallenge} -- LQI and IVE.  The final category is inspired by our inspection of a random subset of visual questions with their answers -- GRN.  We checked whether this taxonomy provided full coverage of plausible reasons for answer differences by conducting a pilot crowdsourcing study, and found no additional categories were identified\footnote{Using the setup from the next subsection, we crowdsourced five labels per VQA for 100 VQAs randomly selected from \textit{VizWiz}~\cite{gurari2018VizWizGrandChallenge}.  In this user interface, users were provided an ``OTHER'' category and an optional, open-ended feedback comment box.  No additional reasons were identified either by users selecting ``OTHER'' or providing feedback.}.  This taxonomy is divided into reasons that arise from the visual question versus answers.  

\vspace{-0.75em}
\paragraph{Approach for Labeling VQAs.}
We designed our user interface for this annotation task to show the visual question (used interchangeably with Question-Image/QI pair) with the ten answers, and to require a user to select all reasons why the answers differed.  The user was required to select at least one reason.  We grouped reasons based on whether the issue stemmed from the QI pair versus answers to assist the user in deciding which reason(s) to select.  We also included a reason called \textbf{\textit{Other} (OTH)}, linked to a free-entry text-box, so users could suggest what they thought was the relevant reason when they felt no options were suitable.  

We chose to label \num{\VQsCrowdsourced} VQAs coming from two popular VQA datasets.  Each VQA consists of an image and a question paired with 10 answers crowdsourced from Amazon Mechanical Turk (AMT) workers.  We include the \textit{VizWiz}~\cite{gurari2018VizWizGrandChallenge} dataset to address immediate, real-world needs of people who are blind.  It originates from people who are blind who used mobile phones to snap photos and record questions about them~\cite{bigham2010VizWiznearlyrealtime,brady2013visual}; e.g., ``what type of beverage is in this bottle?'' or ``has the milk expired?''.  We used the entire dataset, excluding VQs where all answers are identical with exact string matching (i.e., no answer differences), resulting in \num{\VizWizTotal} VQAs.  We also included a sample from the popular \textit{VQA\_2.0}~\cite{goyal2017MakingVQAmatter} dataset for comparison.  It differs from \textit{VizWiz} in part because the images and questions were created separately.  The images originate from the MS-COCO dataset \cite{lin2014MicrosoftCOCOCommon} and the questions came from crowd workers instructed to ask a question about the image that can `stump' a `smart robot'~\cite{antol2015VqaVisualquestion}.
We used a subset of \num{\VQATotal} randomly selected QI pairs from the training set, for which the ten crowdsourced answers were not identical using exact string matching.  Together, VizWiz~\cite{gurari2018VizWizGrandChallenge} and VQA\_2.0~\cite{goyal2017MakingVQAmatter} represent a diversity of users and use case scenarios.  

We employed crowd workers from Amazon Mechanical Turk to label the VQAs.  For quality control, we restricted workers to those who previously completed over 500 jobs with at least a 95\% approval rating, and are from the US to try to ensure English proficiency.  We also collected five labels per visual question from five workers.

%% file: dataset-analysis.tex
\begin{figure}[!b]
\centering
\includegraphics[clip=true, trim=0 235 618 0, width=\linewidth]{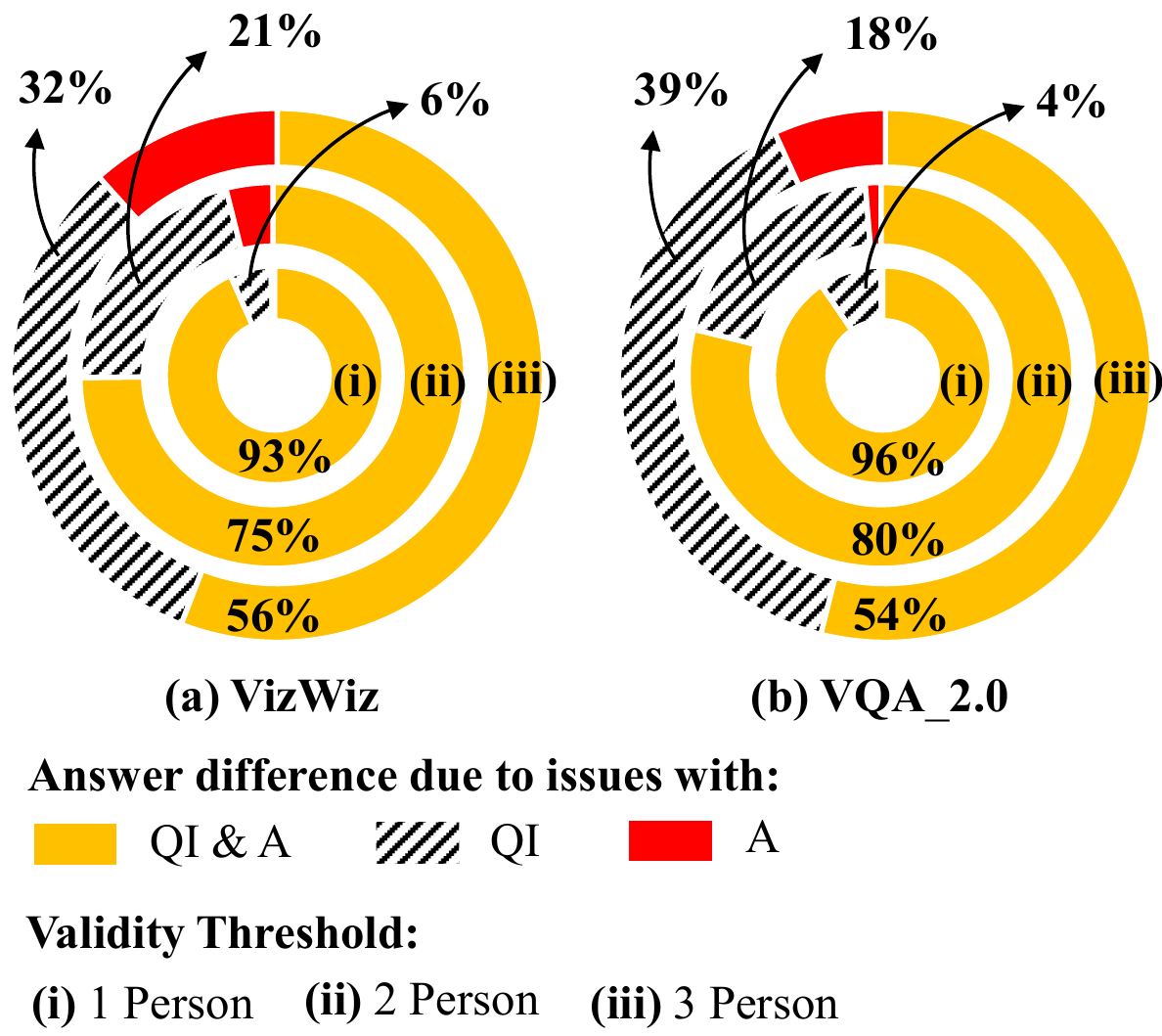} 
\vspace{-0.75em}
\caption{Percentage of VQAs where answer differences arise due to issues with both the QI pair and the 10 answers (\textbf{QI \& A}, yellow), issues with the QI pair only (\textbf{QI}, striped), or issues with the 10 answers only (\textbf{A}, red), for the \textbf{(a)} \textit{VizWiz} and \textbf{(b)} \textit{VQA\_2.0} datasets.  Results are shown with respect to different levels of trust in the crowd workers: \textbf{(i)} \textit{Trust All}: only one worker has to select the reason (1 person validity threshold); \textbf{(ii)} \textit{Trust Any Pair}: at least two workers must agree on the reason (2 person validity threshold); and \textbf{(iii)} \textit{Trust Majority}: at least three workers must agree on the reason (3 person validity threshold).}
\label{fig:dis_src_typ}
\end{figure}

\begin{figure*}[t!]
\captionsetup[subfloat]{farskip=0pt,captionskip=0pt}
\centering
\includegraphics[clip=true, trim=0 210 0 0, width=\linewidth]{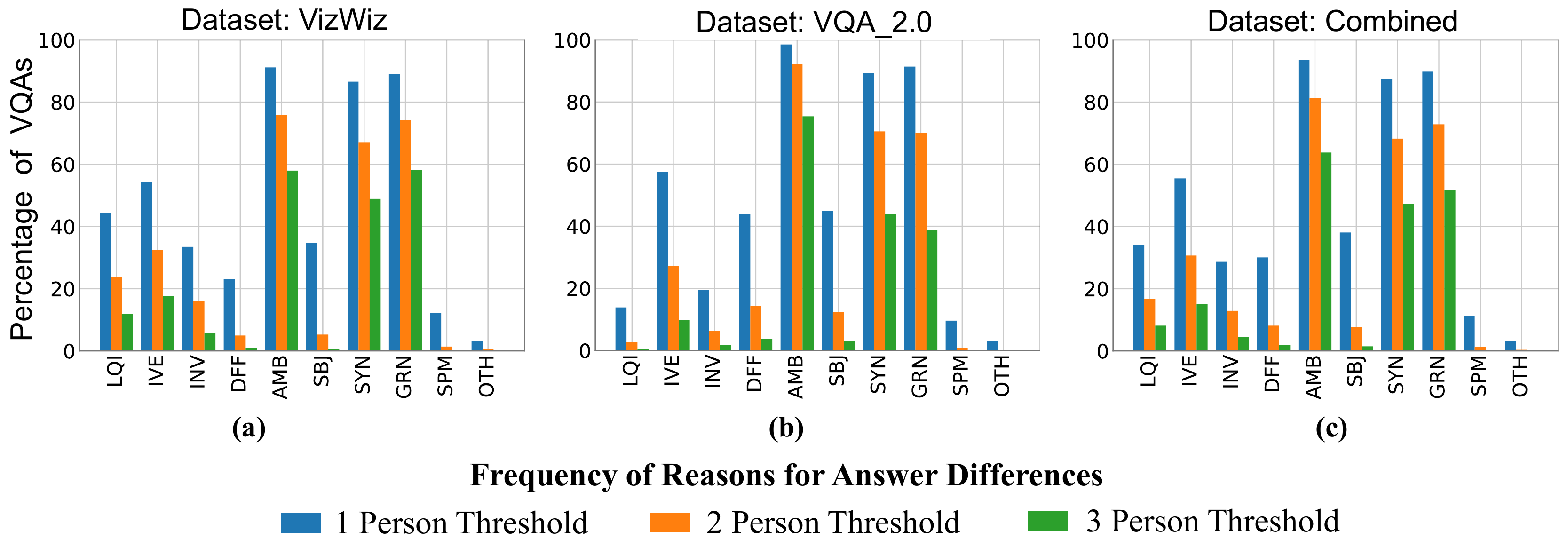} 
\caption{Histograms showing the frequency of each reason leading to answer differences for (a) \num{\VizWizTotal} visual questions asked by blind people, (b) \num{\VQATotal} visual questions asked by sighted people, and (c) combination of the two.  The plots are computed based on increasing thresholds of inter-worker agreement required to make a reason valid ranging from requiring at least one worker selecting it up to at least three workers.  The most popular reasons are ambiguous visual questions (AMB), synonymous answers (SYN), and varying answer granularity (GRN) whereas the most rare are spam (SPM) and other (OTH).}
\label{fig:hist_lbl_freq}
\end{figure*}

\section{Understanding Why Answers Differ}
\label{sec:descriptive}
We analyzed the \num{\AnnsCrowdsourced} crowdsourced annotations to answer the following: (1) what reasons (un)commonly evoke differing answers?, (2) how many unique reasons typically lead to differing answers for a visual question?, and (3) which reasons commonly occur together versus alone?

\subsection{(Un)Common Reasons for Answer Differences}
\label{sec:lbl_freq}

\paragraph{QI Pair Versus Answers.}  We first quantified the tendency for answer differences to arise because of issues with the QI pair versus answers.  To do so, we quantified the proportion of VQAs having issues only with the QI pair (i.e., due to LQI, IVE, INV, DFF, AMB, or SBJ), only with the answers (i.e., due to SYN, GRN, or SPM), and with both.  To enrich our analysis, we examined the influence of different levels of trust in the crowd by tallying the valid reasons observed when requiring a minimum of 1, 2, or 3 members of the crowd to offer the same reason for the reason to be valid.  Results for each dataset are shown in Figure~\ref{fig:dis_src_typ}.  

We found that answer differences arise most often because of issues with both the QI pair and the answers. 
For example, 75\% and 80\% of visual questions from \textit{VizWiz} and \textit{VQA\_2.0} datasets, respectively, have issues arising from both sources (Figure~\ref{fig:dis_src_typ}; 2 person threshold).  In contrast, a very small fraction of answer differences arise because of the answers alone.  A larger percentage arise because of issues with the QI pair alone, affecting roughly 20\% of VQAs for both datasets (Figure~\ref{fig:dis_src_typ}; 2 person threshold).  This latter finding highlights that a considerable portion of answer differences could be avoided by modifying the visual question.

\vspace{-0.75em}
\paragraph{Frequency of Each Reason.}
We next examined the tendency for each of the nine reasons to lead to different answers.  To do so, we calculated the percentage of VQAs assigned with each reason for different levels of trust in the crowd (i.e., requiring a minimum of 1, 2, or 3 members of the crowd to offer the same reason for the reason to be valid).  Results are shown in Figure~\ref{fig:hist_lbl_freq}.

The most common reasons match for both datasets: ambiguous QIs (AMB), followed by synonymous answers (SYN), and finally varying levels of answer granularity (GRN). 
Ambiguity (AMB) accounts for 81.3\% of answer disagreements across both datasets (Figure~\ref{fig:hist_lbl_freq}c; 2 person threshold).  
Ambiguous examples in the \textit{VizWiz} dataset often arise because the question ``What (object) is this \ldots ?'' is asked about an image showing multiple objects (e.g., `store', `shopping area', `shopping cart').
In the \textit{VQA\_2.0} dataset, we found ambiguity can arise for lengthy questions which may leave individuals confused about how to interpret the question (\eg `What weather related event can be seen under the clouds in the horizon?') as well as for visual questions seemingly designed to be ambiguous in order to ``stump a robot"~\cite{antol2015VqaVisualquestion} (\eg `Q: Where are the baby elephants? Ans 1: right, Ans 2: on the grass, Ans 3: next to mom and dad, etc.).  The closely-following second and third most common reasons are answer granularity (GRN) and synonyms (SYN) which account for 72.9\% and 68.3\% of VQAs across both datasets (Figure~\ref{fig:hist_lbl_freq}c; 2 person threshold).  These findings highlight that most answer differences can be resolved by disambiguating visual questions or resolving synonyms and differing granularity~\cite{gadiraju2017Clarityworthwhilequality,li2017LearningDisambiguateAsking,wan2016ModelingAmbiguitySubjectivity}.   

\begin{figure*}[!t]
\centering
\includegraphics[clip=true, trim=0 300 132 0, width=\linewidth]{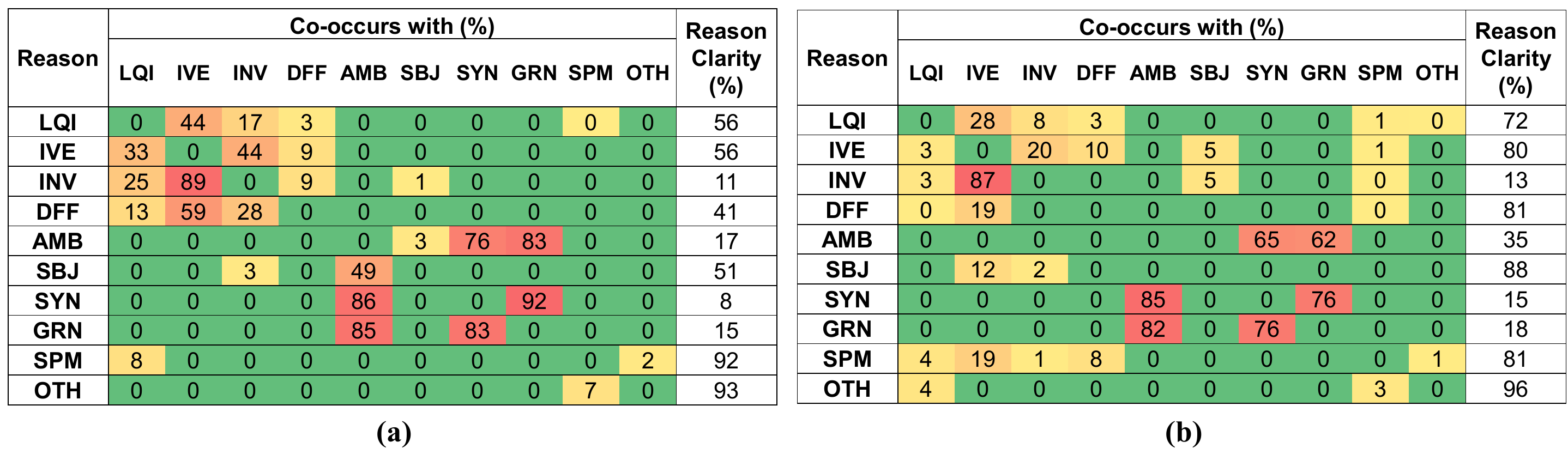} 
\vspace{-1.5em}
\caption{Tendency for each reason to co-occur with every other reason as well as to occur alone (i.e., reason clarity) for the (a) \textit{VizWiz} and (b) \textit{VQA\_2.0} datasets (2-person threshold).  Across both datasets, `spam' and `other' most often occur alone.}
\label{fig:d_sim_amb}
\end{figure*}

The least common reason for answer differences is \textbf{spam (SPM)}, with it accounting for approximately 1.1\% of VQAs across both datasets (Figure~\ref{fig:hist_lbl_freq}c; 2 person threshold).  This is interesting because the issue of spam has received a lot of attention in the crowdsourcing literature (e.g., \cite{eickhoff2013Increasingcheatrobustness, gadiraju2015Understandingmaliciousbehavior,vuurens2012obtaining} to name a few).  Our findings suggest that improving spam detection solutions will lead to considerably less impact than improving approaches addressing the other reasons.

Despite that the approaches for curating the VQAs in the \textit{VizWiz} and \textit{VQA\_2.0} datasets are very different---with \textit{VizWiz} arising from daily visual challenges of blind users, and \textit{VQA\_2.0} containing visual questions designed to be hard for machines to answer---we found overall that the ranking and prevalence of reasons for answer difference across the two datasets is very similar.  
The key differences lie in difficult VQs (DFF) and low quality images (LQI). 
For example, the percentage of difficult visual questions is four times more in \textit{VQA\_2.0} than \textit{VizWiz}; i.e., $\sim$3\% versus $\sim$12\% (2-person threshold).  Additionally, the percentage of low quality images in \textit{VizWiz} accounts for approximately nine times more than that observed for \textit{VQA\_2.0}; i.e., 23.8\% versus 2.6\% of all visual questions respectively (2-person threshold).  Despite such dataset differences, we will show in Section~\ref{sec:prediction_system} that prediction models can still learn to predict which reason(s) will lead to answer differences.

Examples of VQAs that were most confidently voted by the crowd workers as belonging to each of the nine reasons for answer differences are shown in Figure~\ref{fig:disagreement_examples}.  

\subsection{Co-Occurring Reasons for Answer Differences}
\label{sec:unq_lbl}
We now examine how many reasons typically lead to answer differences for a given VQA as well as to what extent reasons co-occur.  For the following analysis, we assume a reason occurs for a VQA if at least two crowd workers flag that reason as occurring.  

\vspace{-0.75em}
\paragraph{Number of Unique Reasons.} We first tallied the number of unique reasons leading to answer differences for each VQA.  Across both datasets, there are most commonly three unique reasons; i.e., for more than 55\% of the \textit{VizWiz} and \textit{VQA\_2.0} VQAs.  Two and four reasons also are common, accounting for 15\% and 16\% VQAs respectively across both datasets.  The remaining $\sim$20\% arise from one unique reason, followed by five and six unique reasons respectively, across both datasets.  These findings motivate representing the problem of predicting which reasons lead to answer differences as a multi-label classification problem.

\vspace{-0.75em}
\paragraph{Reasons Occurring Together.}
We next examined the extent to which the various reasons co-occur.  To do so, we computed an adaptation of causal power \cite{cheng1997covariation,luhmann2005meaning}, measuring the co-occurrence of two reasons $d_i$ and $d_j$ as:
\vspace{-0.5em}

\begin{equation} \label{eqn:d_co_occur}
co\_occurrence ~ ({d_i},{d_j}) = {{P({d_j}|{d_i}) - P({d_j}|{{\bar d}_i})} \over {1 - P({d_j}|{{\bar d}_i})}}    
\end{equation}

\noindent
where $P(d)$ is the probability that reason $d$ is present for a VQA and $P({\bar d})$ is the probability that it is not present.  Intuitively, this metric indicates how often $d_j$ arises when $d_i$ occurs for a VQA.  Results are shown in Figure~\ref{fig:d_sim_amb}.

Across both \textit{VizWiz} and \textit{VQA\_2.0}, we observe the reasons with the highest co-occurrences ($\geq$ 80\%) are answer synonyms (SYN), answer granularity (GRN), invalid questions (INV), and ambiguity (AMB). 
For example, in \textit{VizWiz}, for all the VQAs where SYN was chosen, GRN co-occurs for 92\% of those VQAs, followed by AMB for 86\%.  Likewise, in all the question where GRN occurs, AMB occurs in 85\% of them, and SYN occurs in 83\% of them.  Our results offer strong evidence that ambiguity in the QI pair can cause people to be uncertain both about what level of detail to provide and what word to use among valid synonyms; e.g., `money', `currency', `10 dollar bill'.  We hypothesize that one promising way a VQA system could greatly reduce the frequency of answer differences is to instruct individuals asking QIs to clarify the level of detail they are seeking whenever answer differences are expected to arise from QI ambiguity or differing answer granularity.  

We also observe that across both datasets, invalid question (INV) is paired with insufficient visual evidence to answer the question (IVE).  This is true for 89\% of VQAs in \textit{VizWiz} and 87\% of VQAs in \textit{VQA\_2.0}.  This suggests that if the answer to the question is not present in the image, people think that the question is invalid.

\begin{figure*}[t!]
\centering
\includegraphics[width=0.85\textwidth]{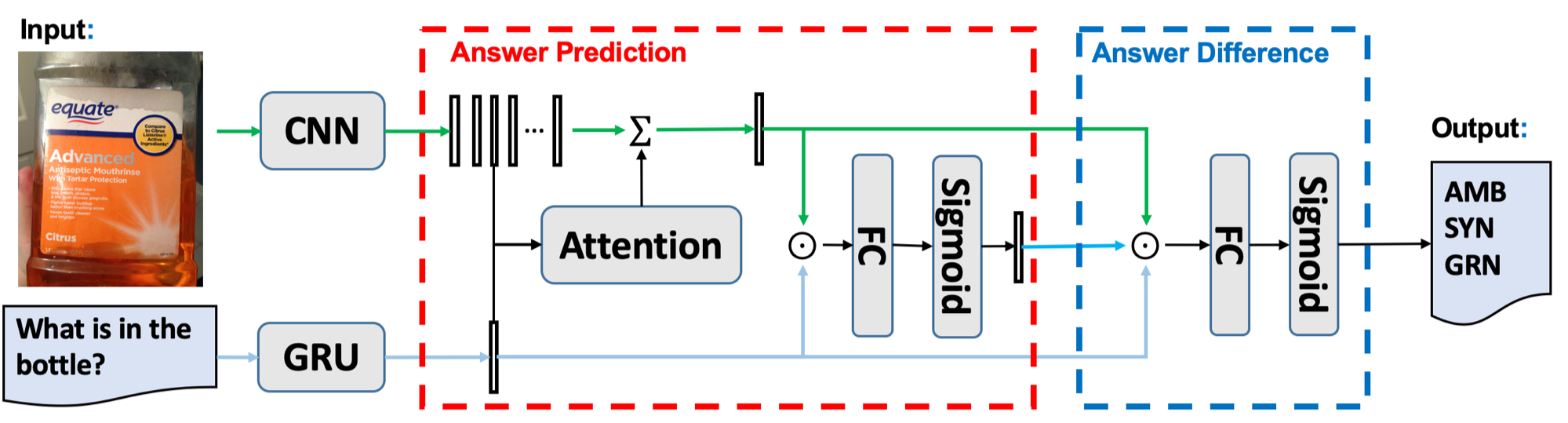}
\caption{Summary of the proposed model (Q+I+A) for predicting which among 10 reasons will lead to answer differences.}
\label{fig:framework}
\end{figure*}

\vspace{-0.75em}
\paragraph{Reasons Occurring Alone.}
We next measured how often a reason occurs on its own.  To do so, we compute \textit{clarity} of a reason $d$ as follows: the percentage of all VQAs where $d$ is chosen for which no other reason is chosen.  Results are shown in Figure~\ref{fig:d_sim_amb}.  Across both datasets, spam (SPM) commonly occurs alone; i.e., for at least 92\% and 81\% of the VQAs in \textit{VizWiz} and \textit{VQA\_2.0}, in which this reason is valid.  For \textit{VizWiz}, as reflected by small percentage values for clarity, most of the reasons commonly co-occur with at least one other label.  In contrast, \textit{VQA\_2.0} only has four of the nine labels (INV, SYN, GRN, AMB) with low clarity ($\leq 35\%$), while the rest have high clarity ($\geq 72\%$).  This reveals that co-occurrence in \textit{VQA\_2.0} centers around fewer reasons than for \textit{VizWiz}.

%% file: prediction-system.tex
\section{Predicting Why Answers Will Differ}
\label{sec:prediction_system}
We now introduce a novel machine learning task of predicting why a visual question will lead to different answers.  

\subsection{Prediction Model}
\label{sec:ml_setup}
We pose the task as a multi-label classification problem.

\vspace{-0.75em}
\paragraph{Ground Truth.}
We compute binary ground truth for each of the nine reasons described in Table~\ref{tab:tab_taxonomy}, as well as the ``Other" category.  Consequently, our ground truth consists of 10 labels.  For each label,  we consider it as present (i.e., `1') for a visual question only if at least two of the five crowd workers selected that label as present.

\vspace{-0.75em}
\paragraph{Proposed Model.}
The motivation for the design of our model is that it employs as predictive cues the input image (I), input question (Q), \emph{and} answers to the QI pair (A).  This is because we know that answer differences can arise from the QI pair alone (i.e., first six reasons in Table~\ref{tab:tab_taxonomy}) as well as from the answers alone (i.e., last three reasons in Table~\ref{tab:tab_taxonomy}).  Our key challenge is how to represent the answers, since the ground truth answers are not known in practice.

Our model is summarized in Figure~\ref{fig:framework}.  It takes as input the \emph{image}, encoded as the last convolutional layer of the Faster R-CNN model~\cite{ren2015faster}, and the \emph{question}, encoded as a 300 dimension word from the pre-trained GloVe vectors~\cite{pennington2014glove} that is then passed to a single-layer GRU with 1024 hidden units.  Our key design decision was to pass this input to an ``Answer Prediction" module, with the goal that it outputs all answers.  We employed a VQA algorithm~\cite{Anderson2017up-down,Teney2018TipsAT} that is trained to optimize for multiple correct answers and output a ``soft target" vector representation of its confidence on all answer candidates; e.g., for the second example in second row of Figure~\ref{fig:disagreement_examples}, the ideal prediction for the 10 answers of 7 ``pillows", 2 ``blanket", and 1 ``sheets" would be 0.7 for ``pillows", 0.2 for ``blanket", 0.1 for ``sheets", and 0 otherwise.\footnote{To avoid learning on the same data observed at test time, we ensure the ``Answer Prediction" module is pretrained on a different VQA dataset.}  By including the ``Answer Prediction" module, the subsequent ``Answer Difference" module can then make predictions based not only on the image and question features but also using the (predicted) answers.  The ``Answer Difference" module passes its input (Q, I, \& A) through a fully-connected layer of 1024 units to a sigmoid function in order to predict the probability of each of the 10 labels.


For training, we adopt the binary cross entropy loss as:

\vspace{-1em}
\begin{align*}
    L = \sum_{i=1}^N y_{i} \log (p_i) + (1-y_{i}) \log (1 - p_{i}),
\end{align*}

\noindent
where $N$ is the total number of labels, $y_i$ is the ground truth label, and $p_i$ is the predicted probability from the sigmoid function.  We initialize the model using the pre-trained weights of \cite{pennington2014glove} for the ``GRU", pre-trained weights of \cite{Anderson2017up-down} for the ``CNN" and ``Answer Prediction" modules, and random values for the ``Answer Difference" module.  We fine-tune the whole network using the Adam solver with a fixed learning rate of 0.001 and batch size of 128.  Dropout and early stopping (five epochs) are used to reduce overfitting. 

\begin{table*}[htbp]
  \centering
  \caption{Average precision for predicting why answers to visual questions will differ for the VQA\_2.0 and VizWiz datasets.}
  \vspace{-0.5em}
    \begin{tabular}{clrrrrrrrrrrr}
    \toprule
    &\textbf{Model} & \textbf{Overall} & \textbf{LQI} & \textbf{IVE} & \textbf{INV} & \textbf{DFF} & \textbf{AMB} & \textbf{SBJ} & \textbf{SYN} & \textbf{GRN} & \textbf{SPM} & \textbf{OTH} \\
    \midrule
    \multirow{8}{*}{\rotatebox[origin=c]{90}{VQA\_2.0}} &
    Random &	30.24 &	3.71&	22.43&	15.09&	14.62&	95.19&	14.18&	64.99&	69.42&	0.52&	\textbf{2.25} \\
&QI-Relevance~\cite{mahendru2017promise} &	32.23&	4.01&	43.16&	15.09&	14.62&	94.11&	14.18&	64.99&	69.42&	0.52&	2.25 \\
&I&	31.88&	4.31&	29.46&	9.28&	17.02&	92.91&	17.99&	74.55&	72.56&	0.5&	0.24 \\
&Q&	43.47&	7.65&	58.89&	44.56&	28.15&	96.42&	24.04&	88.63&	84.67&	\textbf{1.36}&	0.38 \\
&Q+I&	43.16&	9.05&	58.03&	41.95&	28.22&	96.25&	24.29&	88.26&	84.02&	1.27&	0.26\\
&Q+I+A&	\textbf{44.55}&	\textbf{11.58}&	59.95&	\textbf{46.03}&	30.27&	96.47&	24.88&	89.69&	85.62&	0.8&	0.26\\
&Q+I+A\_FT &	44.46&	8.11&	\textbf{60.67}&	43.36&	\textbf{31.35}&	\textbf{96.98}&	\textbf{25.31}&	\textbf{90.49}&	\textbf{86.89}&	1.03&	0.48\\
&Q+I+A\_GT&	44.09&	8.94&	59.64&	45.21&	30.04&	96.60&	23.83&	89.75&	85.80&	0.82&	0.30\\ \hline \hline
\multirow{9}{*}{\rotatebox[origin=c]{90}{VizWiz}} &
Random           & 30.15          & 23.59          & 33.69          & 18.15          & 5.70           & 74.70          & 5.14           & 66.61          & 71.94          & 1.35          & 0.62          \\
&QI-Relevance~\cite{mahendru2017promise}     & 31.71          & 30.56          & 40.52          & 18.15          & 5.7            & 76.53          & 5.14           & 66.61          & 71.94          & 1.35          & 0.62          \\
&Unanswerable~\cite{gurari2018VizWizGrandChallenge}     & 35.31          & 44.82          & 58.63          & 18.15          & 5.7            & 80.14          & 5.14           & 66.61          & 71.94          & 1.35          & 0.62          \\
&I                & 40.54          & 55.42          & 50.66          & 30.12          & 8.77           & 83.39          & 8.64           & 79.76          & 86.29          & 1.71          & 0.61          \\
&Q                & 40.5           & 35.87          & 54.66          & 39.24          & 12.32          & 84.41          & 11.00          & 79.46          & 85.10          & \textbf{2.15}         & 0.76          \\
&Q+I              & 45.73          & 57.81          & 62.47          & 43.24          & 13.77          & 87.81          & 11.14          & 86.36          & 92.01          & 2.00            & 0.75          \\
&Q+I+A            & \textbf{50.02} & \textbf{65.58} & \textbf{77.42} & 56.54          & \textbf{10.49} & \textbf{89.70} & 11.26          & 90.42          & 95.44          & 1.98 & \textbf{1.31} \\
&Q+I+A\_FT & 50.01          & 64.93          & 77.40          & \textbf{56.78} & 10.10          & 89.48          & \textbf{13.16} & \textbf{90.52} & \textbf{95.50} & 1.84          & 1.28          \\
&Q+I+A\_GT        & 50.68          & 66.25          & 77.71          & 57.20          & 13.55          & 90.01          & 12.46          & 90.53          & 95.51          & 1.96          & 1.62  \\

    \bottomrule
    \end{tabular}%
  \label{tab:AP-VQA}%
\end{table*}%

\subsection{Evaluation}

\paragraph{Dataset Train/Validation/Test Split.}
We used the whole \textit{VizWiz} dataset, including the QI pairs where all answers were identical (i.e., 3\% of the VQAs from the original dataset) so that trained algorithms can work well in the presence of QI pairs that do not lead to answer differences.  Using similar train/validation/test splits from \cite{gurari2018VizWizGrandChallenge}, we have 
\num{\VizWizMLTrain} training (\eval{100*\VizWizMLTrain/\VizWizMLTotal}\%), 
\num{\VizWizMLVal} validation (\eval{100*\VizWizMLVal/\VizWizMLTotal}\%), and 
\num{\VizWizMLTest} test (\eval{100*\VizWizMLTest/\VizWizMLTotal}\%) samples. 
For the \num{\VQATotal} VQs from the \textit{VQA\_2.0} dataset, we introduced a 65/10/25 split which resulted in 
\num{\VQATrain} training, 
\num{\VQAVal} validation, and 
\num{\VQATest} test examples.

\vspace{-0.75em}
\paragraph{Evaluation Metrics.}
We report the average precision for each label and the mean average precision across all labels.

\vspace{-0.75em}
\paragraph{Baselines.}
To our knowledge, no prior work has tried to predict the reason(s) why a visual question will have different answers.  Therefore, we evaluate the benefit of three related baseline methods to reveal the value of re-purposing existing approaches for our new problem.\footnote{To avoid overlap between the training data and our test data, we re-trained all baselines with the test set samples excluded (when needed).}  We include random guessing (i.e., \textbf{Random}), since this is the best a user can achieve today.  We also include an algorithm for the related task of predicting whether a question is relevant to the given image, which we call \textbf{QI-relevance}.  We adapt the pretrained question-image-relevance system~\cite{mahendru2017promise} to predict among the plausible reasons: if the QI pair is predicted as relevant, the ``LQI'', ``IVE'', and ``AMB'' are predicted as 0, and 1 otherwise.  Other labels are predicted randomly since they are unrelated to QI-relevance.  We also include an algorithm that predicts whether a VQ is \textbf{unanswerable} for the \textit{VizWiz} dataset (``Q+I'' in ~\cite{gurari2018VizWizGrandChallenge}). Similar to the QI-relevance baseline, if the VQ is predicted as answerable, the ``LQI'', ``IVE'', and ``AMB'' are predicted as 0, and 1 otherwise.

We also evaluate five variants of our \textbf{Q+I+A} model.  We examine the power of different predictive cues by predicting only from the question-image pair (i.e., \textbf{Q+I}), question (i.e., \textbf{Q}), and image (i.e., \textbf{I}).  We examine fine-tuning the VQA model~\cite{goyal2017MakingVQAmatter,gurari2018VizWizGrandChallenge} on our datasets after replacing the last layer of the pre-trained model to
a fully-connected layer representing the answer difference reasons (i.e., \textbf{Q+I+A\_FT}). Finally, we examine what happens when we use the ground truth (GT) instead of the ``Answer Prediction" module ; i.e., directly use a ``soft target" representation of the GT answers rather than predicted answers (i.e., \textbf{Q+I+A\_GT}).  Following \cite{Teney2018TipsAT}, only answers that appear more than eight times in the training dataset are considered in the answer representation.

\vspace{-0.75em}
\paragraph{Results.}
Results for both datasets are shown in Table~\ref{tab:AP-VQA}.

As observed, the proposed model (i.e., \textbf{Q+I+A}) outperforms the existing baselines by a large margin overall.  For example, across both datasets, the performance gain is more than 12 percentage points compared to the next best baseline of \textbf{QI-relevance}.  Compared to the status quo of \textbf{Random} guessing, the gains are even greater. 
The results reveal that re-purposing existing algorithms for our new problem is inadequate, which motivates the need for new algorithmic frameworks that directly learn the ``answer difference" task.

Our findings also reveal the benefit of different predictive cues.  Compared with \textbf{Q+I}, the \textbf{Q+I+A} improves the performance by 1.4\% and 4.3\% on \textit{VQA\_2.0} and \textit{VizWiz} datasets respectively.  This verifies the effectiveness of adding the predicted answers as one more signal for prediction.  Interestingly, we observe \textbf{I} has a much larger impact for the \textit{VizWiz} dataset than for \textit{VQA\_2.0}.  We attribute this difference largely to its significant advantage for the LQI and IVE categories for \textit{VizWiz}, providing over a 20\% and 30\% gain respectively over \textbf{Random} guessing.

When comparing the performance of \textbf{Q+I+A} with \textbf{Q+I+A\_FT} and \textbf{Q+I+A\_GT}, we observe similar performance.  For \textbf{Q+I+A\_FT}, this could be because the predicted answers have already captured the knowledge encoded in the pre-trained VQA model.  For \textbf{Q+I+A\_GT}, this might arise because the answer probabilities from VQA models offer richer information than human-annotated answers and because uncommon answers are excluded.

Overall, the models perform worst across both datasets for VQs that are subjective ($<26\%$) and difficult ($<32\%$).  This highlights a need for models to learn abstract concepts such as common sense.  Other challenges for the model, despite considerable improvement of our models over related baselines (i.e., by typically over $20\%$), are recognizing invalid questions and low quality images. 

%% file: conclusions.tex
\section{Conclusions}
\label{sec:conclusion}
We proposed a taxonomy of nine reasons why answers to visual questions can differ and a novel problem of predicting why answers will differ.  Our experiments demonstrate the promise of algorithms that predict directly from a visual question for this novel task.  The datasets and code are publicly shared at \texttt{https://vizwiz.org} to facilitate future extensions of this work.  Valuable future work includes user studies to identify how to employ such algorithms to guide users in modifying their visual questions so they arrive at a single answer or to retroactively aggregate answers~\cite{sheshadri2013Squarebenchmarkresearch}.

%% file: supp-materials.tex
\onecolumn
\newpage
\appendix
\noindent {\LARGE \textbf{Appendix}}
\vspace{1em}

This document supplements our methods and results provided in the main paper. 

\section{Crowdsourcing Task (supplements section 3 of the main paper)}

\subsection{User Interface}
\begin{figure*}[!h]
\centering
\subfloat[]{\includegraphics[width=0.49\linewidth]{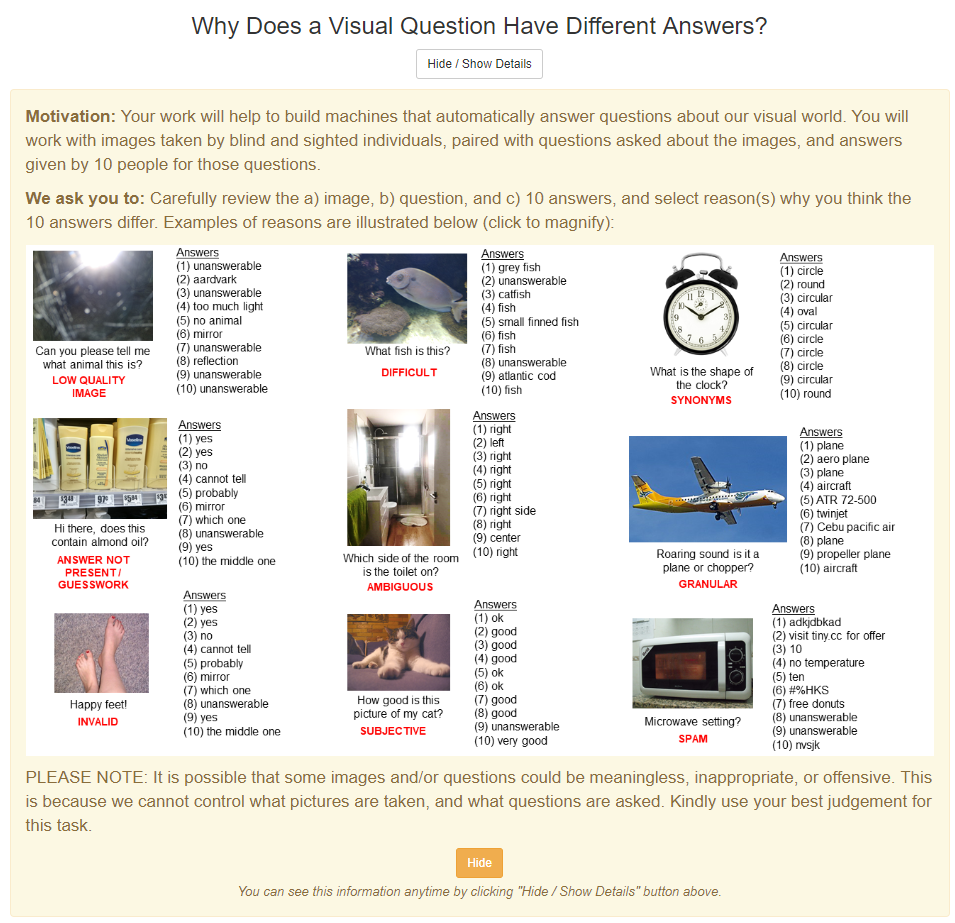} 
\label{fig:task_details}}
\hfill
\subfloat[]{\includegraphics[clip=true, trim=0 0 0 250, width=0.49\linewidth]{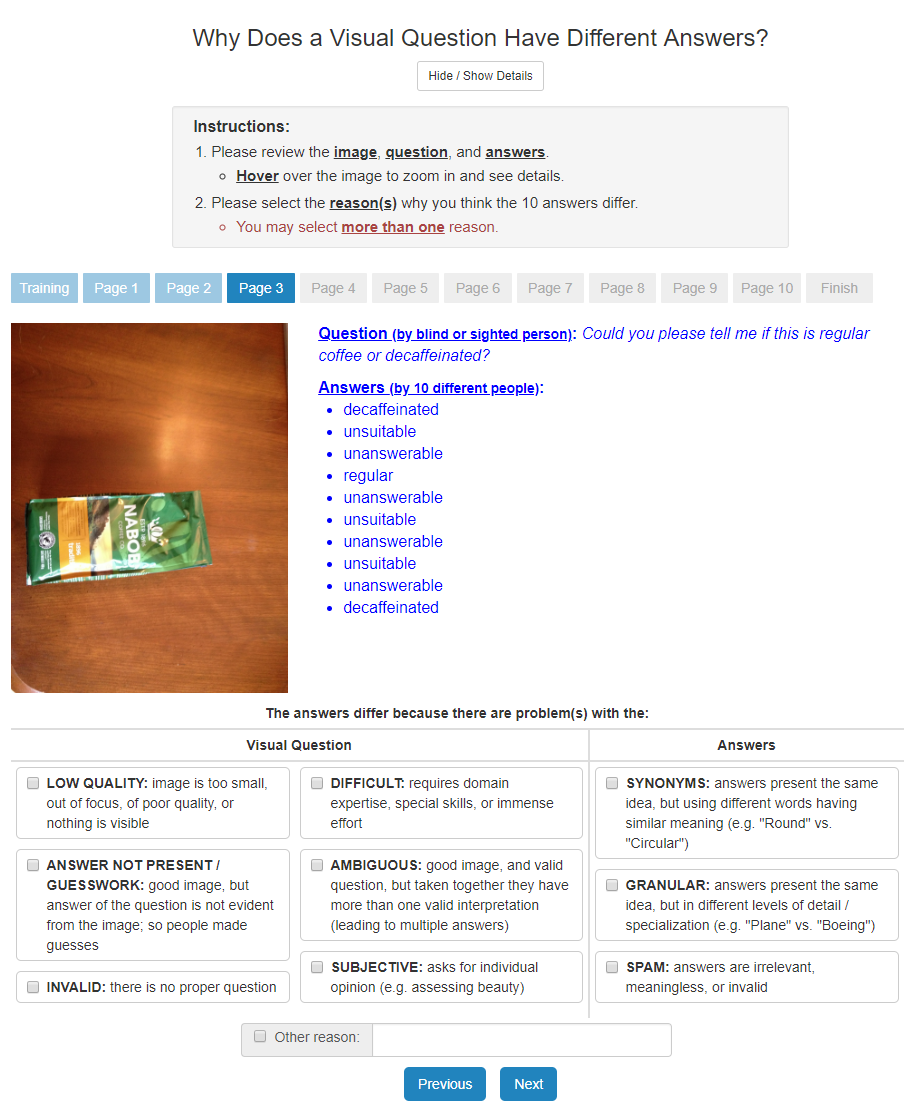} 
\label{fig:task_area}}
\caption{(a) Task instructions to train crowd workers about the reasons that can lead to different answers.
(b) The interface crowd workers used to choose why different answers are observed for a given QI pair with its 10 corresponding answers.}
\label{fig:screenshots}
\end{figure*}

The crowdsourcing user interface is shown in Figure~\ref{fig:screenshots}.

\subsection{Quality Control}
We included a training example that each crowd worker had to complete prior to completing our task.  The authors identified the correct labels beforehand for this example.  For each HIT posted to Amazon Mechanical Turk, the worker had to select these correct labels in order to proceed to the actual task.

\section{Dataset Analysis (supplements section 4 of the main paper)}

\subsection{Inter-Annotator Agreement for Reasons Labels}

We examine inter-annotator agreement among crowd workers.  To do so, we measure the Worker-Worker Similarity (WWS) as the pairwise annotation similarity between two workers across all the VQAs they have annotated in common.  The WWS measure indicates how close a worker performs to the group of workers who have solved the same task.  We calculate WWS between two crowd workers $w_i$ and $w_j$ using three approaches: (a) number of common labels selected, (b) cosine similarity, and (c) Cohen's $\kappa$~\cite{cohen1960coefficient}.

~\\
\textbf{WWS - Common Labels} \\
This metric is defined as 

\[wws({w_i},{w_j}) = \frac{{\sum\limits_{t \in {T_{i,j}}} {numCommonLabels({w_i},{w_j},t)} }}{{\sum\limits_{t \in {T_{i,j}}} {numAnnotations({w_i},t)} }}\]

\noindent
where 
$T_{i,j}$ is the subset of all VQA tasks $T$ annotated by both workers;
$numCommonLabels({w_i},{w_j},t)$  is the number of identical labels selected by both workers $w_i$ and $w_j$ on a VQA task $t$;
and
$numAnnotations({w_i},t)$ is the total number of labels selected by a worker $w_i$ for a single VQA task $t$.

\begin{figure}[!h]
\centering
\includegraphics[clip=true, trim=0 40 0 0, width=0.9\linewidth]{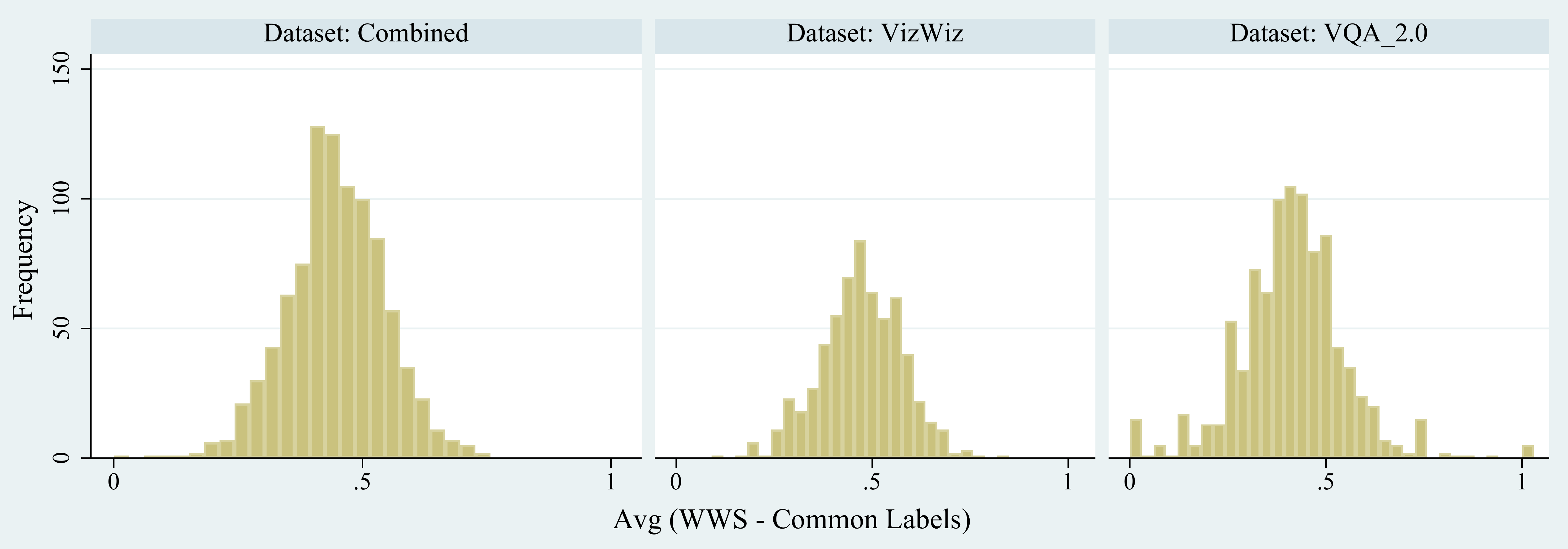} 
\caption{
Distribution of `WWS - Common Labels' for all crowd workers across both datasets alone as well as combined.
}
\label{fig:supp_wws_common_labels}
\end{figure}



\noindent
\textbf{WWS - Cosine Similarity} \\
This metric is defined as follows:

\[avg(\cos ({V_{t,{w_i}}},{V_{t,{w_j}}})) ~~ {\rm{   }}\forall ~~{\rm{  worker }} ~j,{\rm{ }}j \ne i\]

\noindent
where $V_{t,{w_i}}$ is the `Task Vector' of worker $w_i$ annotating VQA task $t$. A Task Vector for a worker annotating a VQA task is defined as a vector whose length is 10 (i.e. equal to the number of labels available), and whose individual elements are either 0 or 1, depending on whether the worker selected the label or not. E.g. if a worker selects the labels LQI, AMB, and SBJ, and the ordering of the labels in the Task Vector are LQI, IVE, INV, DFF, AMB, SBJ, SYN, GRN, SPM, OTH, then the Task Vector becomes:
\texttt{[1,0,0,0,1,1,0,0,0,0]}.

~\\
\textbf{WWS - Cohen's $\kappa$} \\
This metric is defined as follows:
\[ww{s_\kappa } = avg(\kappa )\]

\noindent
where $\kappa$ is the Cohen's kappa coefficient \cite{cohen1960coefficient} used to measure inter-rater agreement.

\begin{figure}[!h]
\centering
\includegraphics[clip=true, trim=0 40 0 0, width=0.9\linewidth]{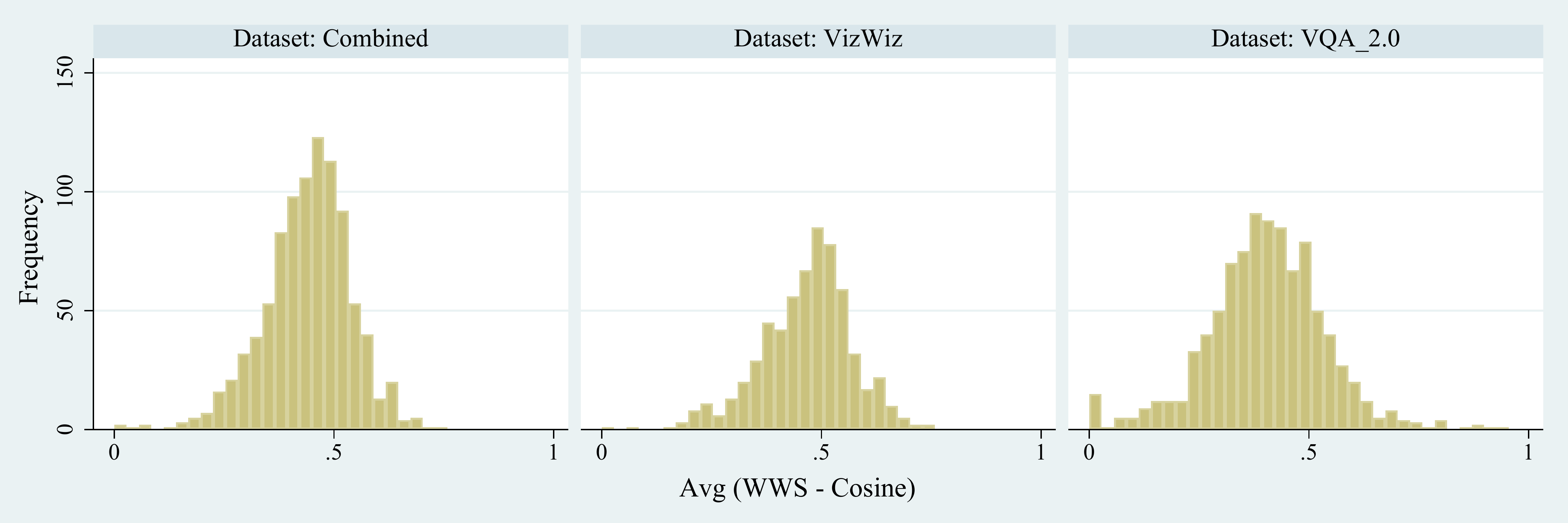} 
\caption{Distribution of `{WWS - Cosine Similarity}' for all crowd workers across both datasets alone as well as combined.}
\label{fig:supp_wws_cosine}
\end{figure}


\begin{figure}[!h]
\centering
\includegraphics[clip=true, trim=0 40 0 0, width=0.9\linewidth]{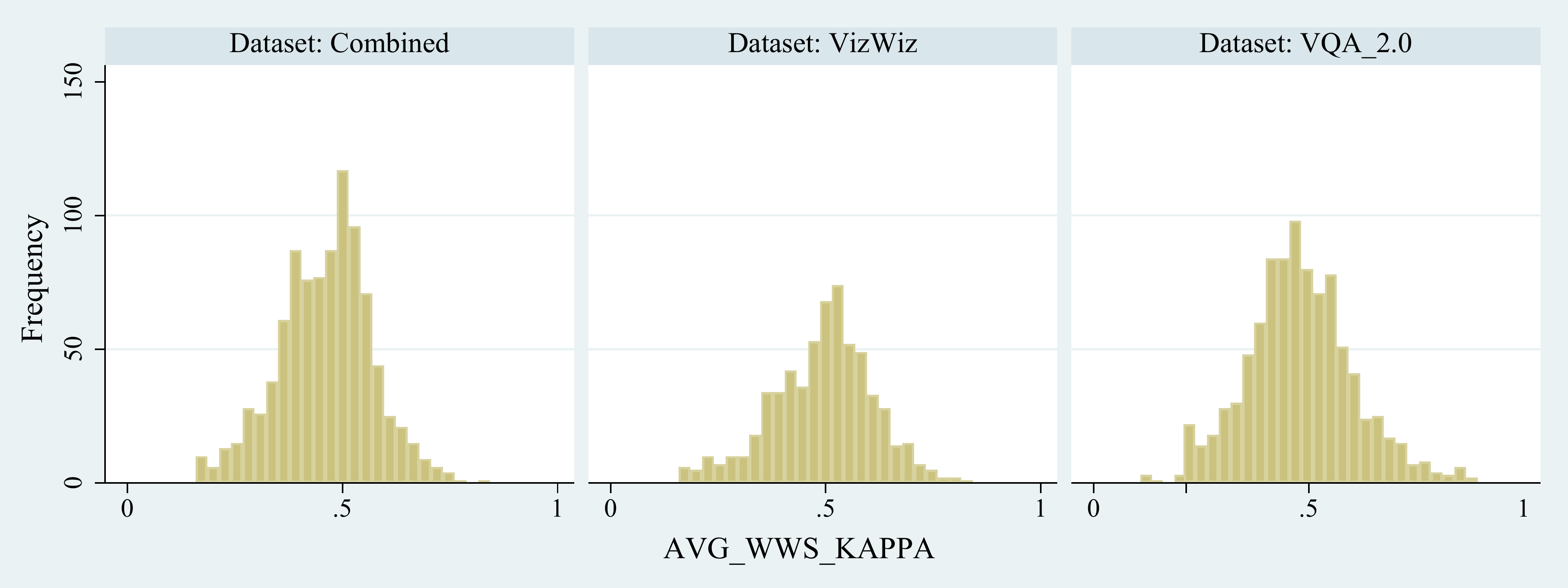} 
\caption{Distribution of `WWS - Cohen's $\kappa$' for all crowd workers across both datasets alone as well as combined.}
\label{fig:supp_wws_kappa}
\end{figure}


Figures \ref{fig:supp_wws_common_labels}, \ref{fig:supp_wws_cosine}, and \ref{fig:supp_wws_kappa} show the distribution of the three WWS metrics for the 934 distinct crowdworkers who provided annotations for our dataset, averaged for each worker.
Among them, 615 distinct workers annotated VQAs from the \textit{VizWiz} dataset, while 928 distinct workers annotated the \textit{VQA\_2.0} dataset.

All the distributions assume an approximately normal form, with peaks at 0.5.  This suggests that most workers agreed with 50\% of the other workers with whom they shared common annotation tasks.

In the case of \textit{VQA\_2.0}, there seems to be a small yet distinct percentage of workers who did not agree with anyone.  This is characterized by a small lump near the 0 value in the plots for \textit{VQA\_2.0}, of all the three WWS metrics (Figures \ref{fig:supp_wws_common_labels}, \ref{fig:supp_wws_cosine}, \& \ref{fig:supp_wws_kappa}).

\subsection{Analysis Using All Validity Thresholds}

\vspace{-1.5em}

\begin{figure}[H]
\centering
\includegraphics[clip=true, trim=0 240 60 0, width=0.95\linewidth]{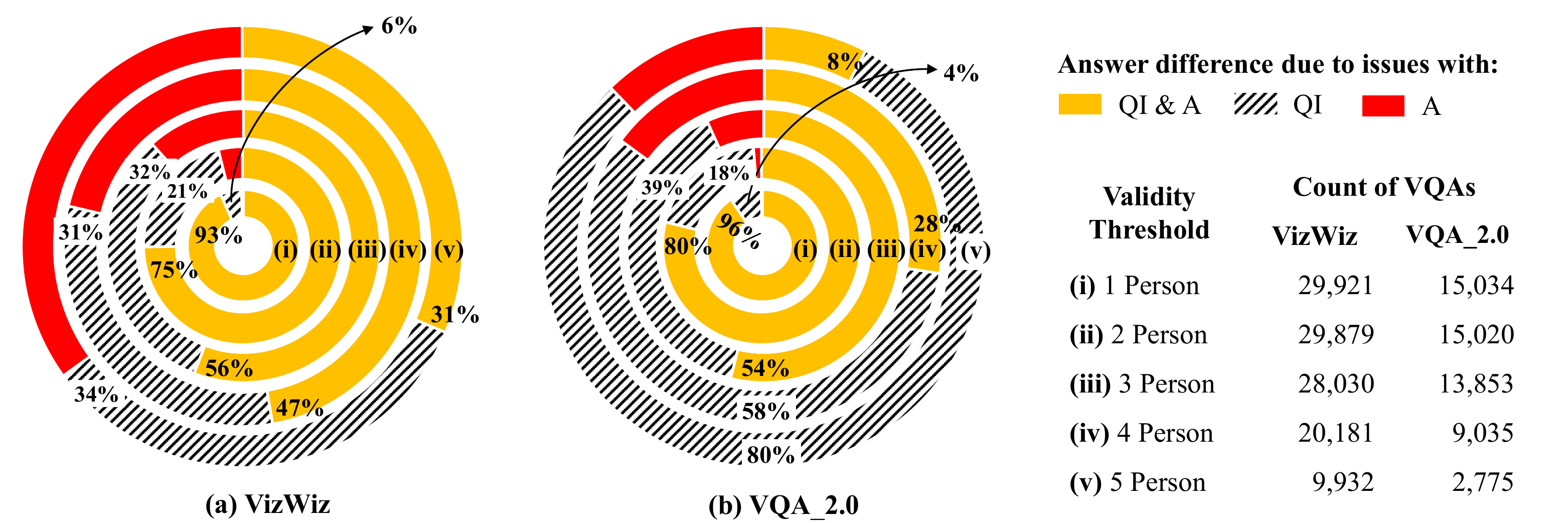} 
\vspace{-0.75em}
\caption{Relative proportion of the various sources of answer disagreement (augmented from Figure 2 in the paper).}
\label{fig:supp_dis_src_typ}
\end{figure}

\vspace{-1.0em}

We tallied the number of reasons leading to answer differences for each VQA, employing various levels of trust in crowd workers: from 1 person threshold to 5 person thresholds.

\begin{figure}[H]
\centering
\includegraphics[clip=true, trim=0 30 0 0, width=0.9\linewidth]{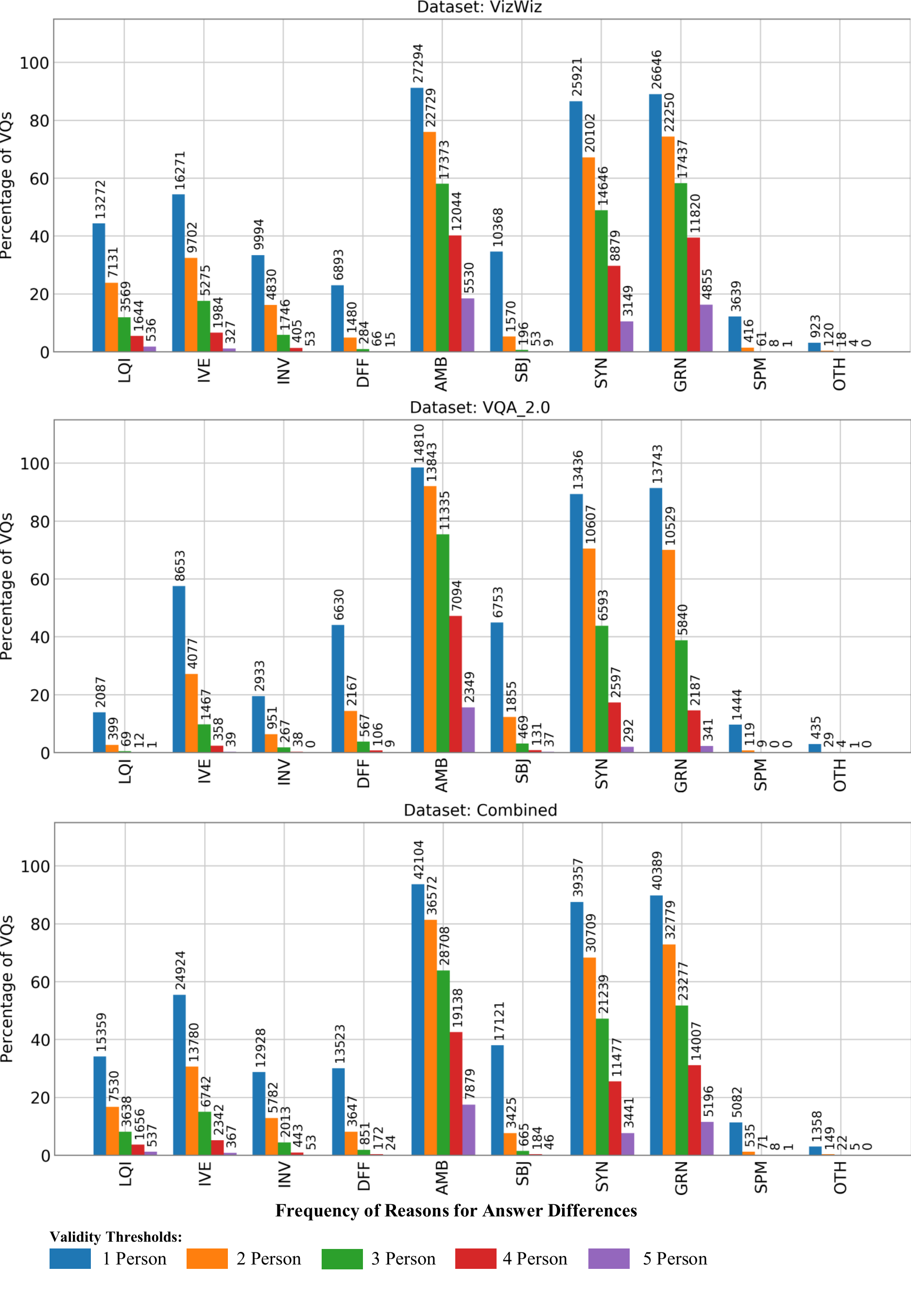} 
\caption{
Histograms showing the frequency of each reason leading to answer differences (augmented from Figure 3 in the paper). 
Data labels show counts of VQAs matching the validity threshold.
}
\label{fig:supp_hist_lbl_freq}
\end{figure}

\begin{figure}[H]
\centering
\includegraphics[clip=true, trim=0 50 0 0, width=0.9\linewidth]{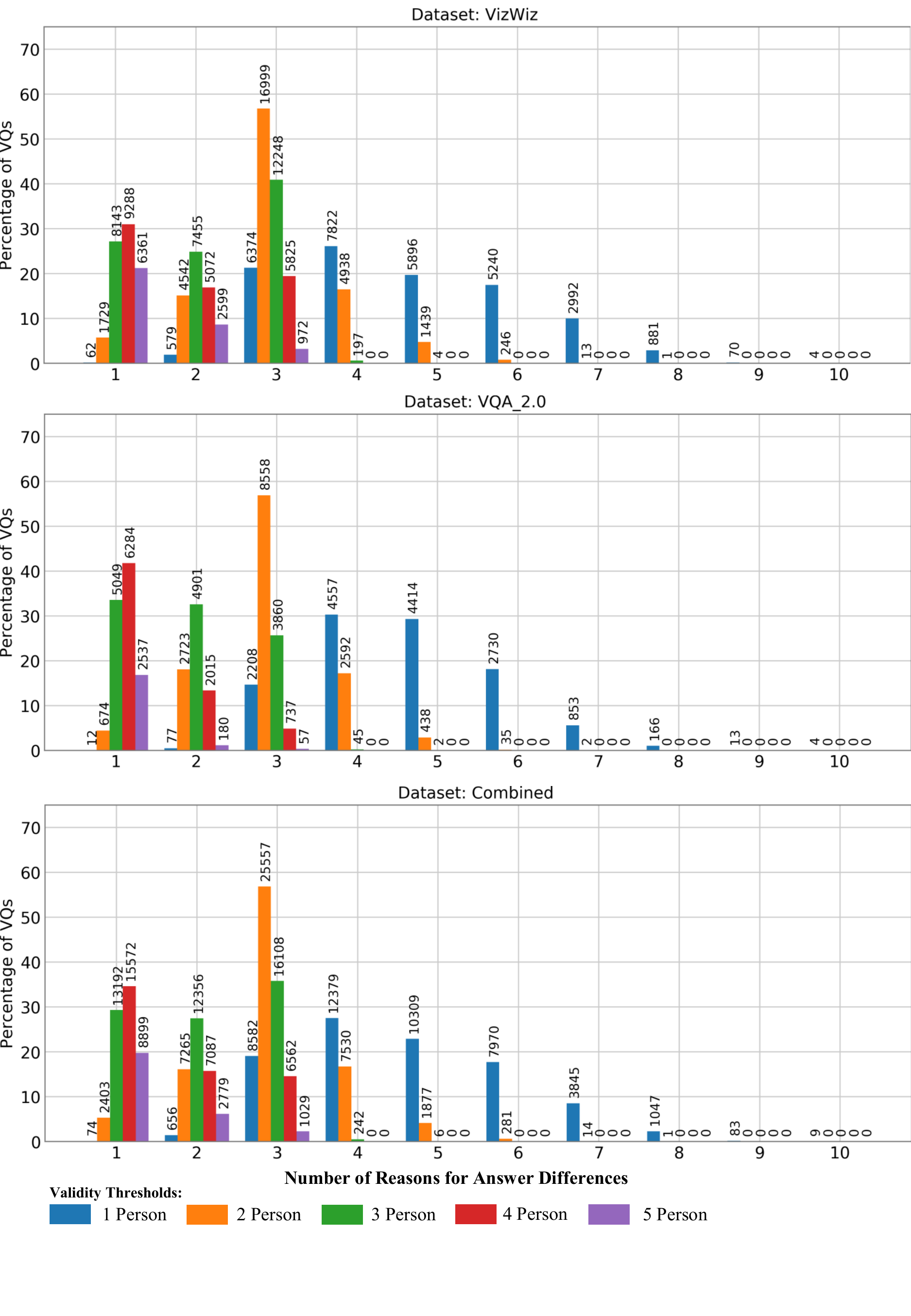} 
\caption{
Histograms showing the number of unique reasons of answer differences identified for each visual question. 
Data labels show counts of VQAs matching the validity threshold.
}
\label{fig:supp_hist_unq_lbl}
\end{figure}

Figure \ref{fig:supp_dis_src_typ} shows the percentage of visual questions, where answer differences arise due to issues with both the QI pair and the 10 answers (\textbf{QI \& A}, yellow), issues with the QI pair only (\textbf{QI}, striped), or issues with the 10 answers only (\textbf{A}, red), for the \textbf{(a)} \textit{VizWiz} and \textbf{(b)} \textit{VQA\_2.0} datasets.  Results are shown with respect to different levels of trust in the crowd workers:  \textbf{(i)} \textit{Trust All}: only one worker has to select the reason (1 person validity threshold); \textbf{(ii)} \textit{Trust Any Pair}: at least two workers must agree on the reason (2 person validity threshold);  \textbf{(iii)} \textit{Trust Majority}: at least three workers must agree on the reason (3 person validity threshold); \textbf{(iv)}: at least four workers must have to select the reason (4 person validity threshold); and \textbf{(v)} \textit{Trust Consensus}: all five workers must agree on the reason (5 person validity threshold).

Figure \ref{fig:supp_hist_lbl_freq} shows histograms of the frequency of each reason leading to answer differences for (a) \num{\VizWizTotal} visual questions asked by blind people (\textit{VizWiz}), (b) \num{\VQATotal} visual questions asked by sighted people (\textit{VQA\_2.0}), and (c) combination of the previous two. The plots are computed based on increasing thresholds of inter-worker agreement required to make a reason valid, ranging from requiring only one worker selecting it (1 person validity threshold) up to all workers agreeing (5 person threshold). The most popular reasons are ambiguous visual questions (AMB), synonymous answers (SYN), and varying answer granularity (GRN) whereas the most rare are spam (SPM) and other (OTH).

Figure \ref{fig:supp_hist_unq_lbl} shows the summary of how many unique reasons are identified as the sources of answer differences for \num{\VizWizTotal} VQs asked by blind people (\textit{VizWiz}), \num{\VQATotal} VQs asked by sighted people (\textit{VQA\_2.0}), and their combination. Across both datasets, most commonly there are three unique reasons for answer differences. Visual inspections show that these are the three most popular reasons: `ambiguous', `synonyms', and `granularity'.

\section{Prediction Model Analysis (supplements section 5 of the main paper)}

\begin{table}[htbp]
  \centering
  \caption{Average precision for predicting why answers to visual questions will differ for the VQA\_2.0 and VizWiz datasets when we exclude the ``spam'' category for training the models.}
    \begin{tabular}{clrrrrrrrrrr}
    \toprule
          & \textbf{Model} & \textbf{Overall} & \textbf{LQI} & \textbf{IVE} & \textbf{INV} & \textbf{DEF} & \textbf{AMB} & \textbf{SBJ} & \textbf{SYN} & \textbf{GRN} & \textbf{OTH} \\
    \midrule
    \multirow{8}{*}{\rotatebox[origin=c]{90}{VQA\_2.0}} & Random & 33.54 & 3.71  & 22.43 & 15.09 & 14.62 & 95.19 & 14.18 & 64.99 & 69.42 & \textbf{2.25} \\
          & QI-Relevance~\cite{mahendru2017promise} & 35.76 & 4.01  & 43.16 & 15.09 & 14.62 & 94.11 & 14.18 & 64.99 & 69.42 & 2.25 \\
          & I     & 35.38 & 3.66  & 29.55 & 10.06 & 17.04 & 93.04 & 18.29 & 74.50 & 72.09 & 0.18 \\
          & Q     & 48.11 & 7.91  & 59.23 & 43.43 & 28.02 & \textbf{96.70} & 23.69 & 88.85 & 84.78 & 0.36 \\
          & Q+I   & 47.87 & 9.18  & 58.83 & 40.83 & 28.18 & 96.48 & 24.20 & 88.39 & 84.37 & 0.37 \\
          & Q+I+A & 48.05 & 7.09  & 59.46 & \textbf{45.18} & 27.99 & 96.60 & 21.89 & 88.97 & 85.05 & 0.26 \\
          & Q+I+A\_FT & \textbf{48.89} & \textbf{9.11} & \textbf{59.78} & 44.99 & \textbf{30.11} & 96.52 & \textbf{24.33} & \textbf{89.49} & \textbf{85.39} & 0.25 \\
          & Q+I+A\_GT & 48.96 & 8.65  & 60.30 & 46.00 & 28.91 & 96.63 & 24.13 & 89.73 & 86.00 & 0.26 \\ \hline \hline
    \multirow{9}{*}{\rotatebox[origin=c]{90}{VizWiz}} & Random & 33.35 & 23.59 & 33.69 & 18.15 & 5.70  & 74.70 & 5.14  & 66.61 & 71.94 & 0.62 \\
          & QI-Relevance~\cite{mahendru2017promise} & 38.76 & 30.56 & 40.52 & 18.15 &  5.7  & 76.53 & 5.14  & 66.61 & 71.94 & 0.62 \\
          & Unanswerable~\cite{gurari2018VizWizGrandChallenge} & 43.26 & 44.82 & 58.63 & 18.15 &  5.7  & 80.14 & 5.14  & 66.61 & 71.94 & 0.62 \\
          & I     & 44.79 & 55.23 & 50.38 & 29.85 & 8.17  & 83.42 & 9.19  & 79.96 & 86.34 & 0.62 \\
          & Q     & 44.76 & 35.38 & 54.43 & 38.91 & 13.59 & 84.44 & 10.59 & 79.68 & 85.15 & 0.65 \\
          & Q+I   & 50.59 & 56.54 & 61.91 & 45.25 & 13.80 & 87.55 & \textbf{11.55} & 85.97 & 91.42 & \textbf{1.36} \\
          & Q+I+A & 55.18 & \textbf{65.51} & \textbf{77.36} & \textbf{55.76} & 10.38 & \textbf{89.77} & 10.83 & \textbf{90.39} & \textbf{95.50} & 1.14 \\
          & Q+I+A\_FT & \textbf{55.35} & 65.30 & 77.18 & 54.19 & \textbf{14.24} & 89.60 & 11.32 & 89.99 & 95.24 & 1.07 \\
          & Q+I+A\_GT & 55.97 & 66.03 & 77.80 & 56.55 & 12.94 & 90.03 & 12.51 & 90.41 & 95.51 & 1.97 \\
    \bottomrule
    \end{tabular}%
  \label{tab:addlabel}%
\end{table}

%% file: ms.bbl
\begin{thebibliography}{10}\itemsep=-1pt

\bibitem{BeSpecular}
{{BeSpecular}}.
\newblock https://www.bespecular.com.

\bibitem{amid2015Multiviewtripletembedding}
Ehsan Amid and Antti Ukkonen.
\newblock Multiview triplet embedding: {{Learning}} attributes in multiple
  maps.
\newblock In {\em International {{Conference}} on {{Machine Learning}}}, pages
  1472--1480, 2015.

\bibitem{amirkhani2014Agreementdisagreementbased}
Hossein Amirkhani and Mohammad Rahmati.
\newblock Agreement/disagreement based crowd labeling.
\newblock {\em Applied intelligence}, 41(1):212--222, 2014.

\bibitem{Anderson2017up-down}
Peter Anderson, Xiaodong He, Chris Buehler, Damien Teney, Mark Johnson, Stephen
  Gould, and Lei Zhang.
\newblock Bottom-up and top-down attention for image captioning and visual
  question answering.
\newblock In {\em CVPR}, 2018.

\bibitem{antol2015VqaVisualquestion}
Stanislaw Antol, Aishwarya Agrawal, Jiasen Lu, Margaret Mitchell, Dhruv Batra,
  C Lawrence~Zitnick, and Devi Parikh.
\newblock {{VQA}}: {{Visual}} question answering.
\newblock In {\em Proceedings of the {{IEEE International Conference}} on
  {{Computer Vision}}}, pages 2425--2433, 2015.

\bibitem{aroyo2013CrowdTruthHarnessing}
Lora Aroyo and Chris Welty.
\newblock Crowd {{Truth}}: {{Harnessing}} disagreement in crowdsourcing a
  relation extraction gold standard.
\newblock {\em WebSci2013. ACM}, 2013, 2013.

\bibitem{bigham2010VizWiznearlyrealtime}
Jeffrey~P Bigham, Chandrika Jayant, Hanjie Ji, Greg Little, Andrew Miller,
  Robert~C Miller, Robin Miller, Aubrey Tatarowicz, Brandyn White, Samual
  White, and {others}.
\newblock {{VizWiz}}: Nearly real-time answers to visual questions.
\newblock In {\em Proceedings of the 23nd Annual {{ACM}} Symposium on {{User}}
  Interface Software and Technology}, pages 333--342. ACM, 2010.

\bibitem{brady2013visual}
Erin Brady, Meredith~Ringel Morris, Yu Zhong, Samuel White, and Jeffrey~P.
  Bigham.
\newblock Visual challenges in the everyday lives of blind people.
\newblock In {\em Proceedings of the SIGCHI Conference on Human Factors in
  Computing Systems}, CHI '13, pages 2117--2126, New York, NY, USA, 2013. ACM.

\bibitem{burton2012crowdsourcing}
Michele~A Burton, Erin Brady, Robin Brewer, Callie Neylan, Jeffrey~P Bigham,
  and Amy Hurst.
\newblock Crowdsourcing subjective fashion advice using vizwiz: challenges and
  opportunities.
\newblock In {\em Proceedings of the 14th international ACM SIGACCESS
  conference on Computers and accessibility}, pages 135--142. ACM, 2012.

\bibitem{chandrasekaran2016Wearehumor}
Arjun Chandrasekaran, Ashwin~K. Vijayakumar, Stanislaw Antol, Mohit Bansal,
  Dhruv Batra, C. Lawrence~Zitnick, and Devi Parikh.
\newblock We are humor beings: {{Understanding}} and predicting visual humor.
\newblock In {\em Proceedings of the {{IEEE Conference}} on {{Computer Vision}}
  and {{Pattern Recognition}}}, pages 4603--4612, 2016.

\bibitem{cheng1997covariation}
Patricia~W Cheng.
\newblock From covariation to causation: a causal power theory.
\newblock {\em Psychological review}, 104(2):367, 1997.

\bibitem{cohen1960coefficient}
Jacob Cohen.
\newblock A coefficient of agreement for nominal scales.
\newblock {\em Educational and psychological measurement}, 20(1):37--46, 1960.

\bibitem{dumitrache2017Crowdsourcinggroundtruth}
Anca Dumitrache, Lora Aroyo, and Chris Welty.
\newblock Crowdsourcing ground truth for medical relation extraction.
\newblock {\em arXiv preprint arXiv:1701.02185}, 2017.

\bibitem{eickhoff2013Increasingcheatrobustness}
Carsten Eickhoff and Arjen~P {de Vries}.
\newblock Increasing cheat robustness of crowdsourcing tasks.
\newblock {\em Information retrieval}, 16(2):121--137, 2013.

\bibitem{gadiraju2015Understandingmaliciousbehavior}
Ujwal Gadiraju, Ricardo Kawase, Stefan Dietze, and Gianluca Demartini.
\newblock Understanding malicious behavior in crowdsourcing platforms: {{The}}
  case of online surveys.
\newblock In {\em Proceedings of the 33rd {{Annual ACM Conference}} on {{Human
  Factors}} in {{Computing Systems}}}, pages 1631--1640. ACM, 2015.

\bibitem{gadiraju2017Clarityworthwhilequality}
Ujwal Gadiraju, Jie Yang, and Alessandro Bozzon.
\newblock Clarity is a worthwhile quality: {{On}} the role of task clarity in
  microtask crowdsourcing.
\newblock In {\em Proceedings of the 28th {{ACM Conference}} on {{Hypertext}}
  and {{Social Media}}}, pages 5--14. ACM, 2017.

\bibitem{goyal2017MakingVQAmatter}
Yash Goyal, Tejas Khot, Douglas Summers-Stay, Dhruv Batra, and Devi Parikh.
\newblock Making the {{V}} in {{VQA}} matter: {{Elevating}} the role of image
  understanding in {{Visual Question Answering}}.
\newblock In {\em {{CVPR}}}, volume~1, page~9, 2017.

\bibitem{gurari2017CrowdVergePredictingIf}
Danna Gurari and Kristen Grauman.
\newblock {{CrowdVerge}}: {{Predicting If People Will Agree}} on the {{Answer}}
  to a {{Visual Question}}.
\newblock In {\em Proceedings of the 2017 {{CHI Conference}} on {{Human
  Factors}} in {{Computing Systems}}}, pages 3511--3522. ACM, 2017.

\bibitem{gurari2018PredictingForegroundObject}
Danna Gurari, Kun He, Bo Xiong, Jianming Zhang, Mehrnoosh Sameki, Suyog~Dutt
  Jain, Stan Sclaroff, Margrit Betke, and Kristen Grauman.
\newblock Predicting {{Foreground Object Ambiguity}} and {{Efficiently
  Crowdsourcing}} the {{Segmentation}} (s).
\newblock {\em International Journal of Computer Vision}, 126(7):714--730,
  2018.

\bibitem{gurari2018VizWizGrandChallenge}
Danna Gurari, Qing Li, Abigale~J. Stangl, Anhong Guo, Chi Lin, Kristen Grauman,
  Jiebo Luo, and Jeffrey~P. Bigham.
\newblock {{VizWiz Grand Challenge}}: {{Answering Visual Questions}} from
  {{Blind People}}.
\newblock In {\em Proceedings of the {{IEEE Conference}} on {{Computer Vision}}
  and {{Pattern Recognition}}}, pages 3608--3617, 2018.

\bibitem{inel2013Domainindependentqualitymeasures}
Oana Inel, Lora Aroyo, Chris Welty, and Robert-Jan Sips.
\newblock Domain-independent quality measures for crowd truth disagreement.
\newblock {\em Detection, Representation, and Exploitation of Events in the
  Semantic Web}, page~2, 2013.

\bibitem{inel2014CrowdtruthMachinehumancomputation}
Oana Inel, Khalid Khamkham, Tatiana Cristea, Anca Dumitrache, Arne Rutjes,
  Jelle {van der Ploeg}, Lukasz Romaszko, Lora Aroyo, and Robert-Jan Sips.
\newblock Crowdtruth: {{Machine}}-human computation framework for harnessing
  disagreement in gathering annotated data.
\newblock In {\em International {{Semantic Web Conference}}}, pages 486--504.
  Springer, 2014.

\bibitem{isola2011Whatmakesimage}
Phillip Isola, Jianxiong Xiao, Antonio Torralba, and Aude Oliva.
\newblock What makes an image memorable?
\newblock In {\em {{CVPR}} 2011}, pages 145--152. {IEEE}, 2011.

\bibitem{jas2015Imagespecificity}
Mainak Jas and Devi Parikh.
\newblock Image specificity.
\newblock In {\em Proceedings of the {{IEEE Conference}} on {{Computer Vision}}
  and {{Pattern Recognition}}}, pages 2727--2736, 2015.

\bibitem{kafle2017analysisvisualquestion}
Kushal Kafle and Christopher Kanan.
\newblock An analysis of visual question answering algorithms.
\newblock In {\em 2017 {{IEEE International Conference}} on {{Computer Vision}}
  ({{ICCV}})}, pages 1983--1991. IEEE, 2017.

\bibitem{kovashka2015Discoveringattributeshades}
Adriana Kovashka and Kristen Grauman.
\newblock Discovering attribute shades of meaning with the crowd.
\newblock {\em International Journal of Computer Vision}, 114(1):56--73, 2015.

\bibitem{li2017LearningDisambiguateAsking}
Yining Li, Chen Huang, Xiaoou Tang, and Chen Change~Loy.
\newblock Learning to {{Disambiguate}} by {{Asking Discriminative Questions}}.
\newblock In {\em 2017 {{IEEE International Conference}} on {{Computer Vision}}
  ({{ICCV}})}, pages 3439--3448, Oct. 2017.
\newblock ISSN:.

\bibitem{lin2014MicrosoftCOCOCommon}
Tsung-Yi Lin, Michael Maire, Serge Belongie, James Hays, Pietro Perona, Deva
  Ramanan, Piotr Doll{\'a}r, and C~Lawrence Zitnick.
\newblock Microsoft {{COCO}}: {{Common}} objects in context.
\newblock In {\em European Conference on Computer Vision}, pages 740--755.
  Springer, 2014.

\bibitem{luhmann2005meaning}
Christian~C Luhmann and Woo-kyoung Ahn.
\newblock The meaning and computation of causal power: Comment on cheng (1997)
  and novick and cheng (2004).
\newblock 2005.

\bibitem{mahendru2017promise}
Aroma Mahendru, Viraj Prabhu, Akrit Mohapatra, Dhruv Batra, and Stefan Lee.
\newblock The promise of premise: Harnessing question premises in visual
  question answering.
\newblock {\em EMNLP}, 2017.

\bibitem{malinowski2015Askyourneurons}
Mateusz Malinowski, Marcus Rohrbach, and Mario Fritz.
\newblock Ask {{Your Neurons}}: {{A Neural}}-{{Based Approach}} to {{Answering
  Questions}} about {{Images}}.
\newblock In {\em 2015 {{IEEE International Conference}} on {{Computer Vision}}
  ({{ICCV}})}, pages 1--9, Dec. 2015.
\newblock ISSN:.

\bibitem{michael2017crowdsourcing}
Julian Michael, Gabriel Stanovsky, Luheng He, Ido Dagan, and Luke Zettlemoyer.
\newblock Crowdsourcing question-answer meaning representations.
\newblock {\em arXiv preprint arXiv:1711.05885}, 2017.

\bibitem{nguyen2016Probabilisticmodelingcrowdsourcing}
An~Thanh Nguyen, Matthew Halpern, Byron~C Wallace, and Matthew Lease.
\newblock Probabilistic modeling for crowdsourcing partially-subjective
  ratings.
\newblock In {\em Fourth {{AAAI Conference}} on {{Human Computation}} and
  {{Crowdsourcing}}}, 2016.

\bibitem{pennington2014glove}
Jeffrey Pennington, Richard Socher, and Christopher Manning.
\newblock Glove: Global vectors for word representation.
\newblock In {\em Proceedings of the 2014 conference on empirical methods in
  natural language processing (EMNLP)}, pages 1532--1543, 2014.

\bibitem{ren2015faster}
Shaoqing Ren, Kaiming He, Ross Girshick, and Jian Sun.
\newblock Faster r-cnn: Towards real-time object detection with region proposal
  networks.
\newblock In {\em Advances in neural information processing systems}, pages
  91--99, 2015.

\bibitem{sharmanska2016AmbiguityHelpsClassification}
Viktoriia Sharmanska, Daniel Hern{\'a}ndez-Lobato, Jose Miguel
  Hernandez-Lobato, and Novi Quadrianto.
\newblock Ambiguity {{Helps}}: {{Classification}} with {{Disagreements}} in
  {{Crowdsourced Annotations}}.
\newblock In {\em 2016 {{IEEE Conference}} on {{Computer Vision}} and {{Pattern
  Recognition}} ({{CVPR}})}, pages 2194--2202, June 2016.
\newblock ISSN:.

\bibitem{sheshadri2013Squarebenchmarkresearch}
Aashish Sheshadri and Matthew Lease.
\newblock Square: {{A}} benchmark for research on computing crowd consensus.
\newblock In {\em First {{AAAI Conference}} on {{Human Computation}} and
  {{Crowdsourcing}}}, 2013.

\bibitem{soberon2013Measuringcrowdtruth}
Guillermo Sober{\'o}n, Lora Aroyo, Chris Welty, Oana Inel, Hui Lin, and Manfred
  Overmeen.
\newblock Measuring crowd truth: {{Disagreement}} metrics combined with worker
  behavior filters.
\newblock In {\em {{CrowdSem}} 2013 {{Workshop}}}, 2013.

\bibitem{teney2018Tipstricksvisual}
Damien Teney, Peter Anderson, Xiaodong He, and Anton {van den Hengel}.
\newblock Tips and tricks for visual question answering: {{Learnings}} from the
  2017 challenge.
\newblock In {\em Proceedings of the {{IEEE Conference}} on {{Computer Vision}}
  and {{Pattern Recognition}}}, pages 4223--4232, 2018.

\bibitem{Teney2018TipsAT}
Damien Teney, Peter Anderson, Xiaodong He, and Anton van~den Hengel.
\newblock Tips and tricks for visual question answering: Learnings from the
  2017 challenge.
\newblock {\em CVPR}, pages 4223--4232, 2018.

\bibitem{vuurens2011much}
Jeroen Vuurens, Arjen~P de Vries, and Carsten Eickhoff.
\newblock How much spam can you take? an analysis of crowdsourcing results to
  increase accuracy.
\newblock In {\em Proc. ACM SIGIR Workshop on Crowdsourcing for Information
  Retrieval (CIR'11)}, pages 21--26, 2011.

\bibitem{vuurens2012obtaining}
Jeroen~BP Vuurens and Arjen~P De~Vries.
\newblock Obtaining high-quality relevance judgments using crowdsourcing.
\newblock {\em IEEE Internet Computing}, 16(5):20--27, 2012.

\bibitem{wan2016ModelingAmbiguitySubjectivity}
Mengting Wan and Julian McAuley.
\newblock Modeling {{Ambiguity}}, {{Subjectivity}}, and {{Diverging
  Viewpoints}} in {{Opinion Question Answering Systems}}.
\newblock In {\em 2016 {{IEEE}} 16th {{International Conference}} on {{Data
  Mining}} ({{ICDM}})}, pages 489--498, Dec. 2016.
\newblock ISSN:.

\bibitem{welinder2010multidimensionalwisdomcrowds}
Peter Welinder, Steve Branson, Pietro Perona, and Serge~J Belongie.
\newblock The multidimensional wisdom of crowds.
\newblock In {\em Advances in Neural Information Processing Systems}, pages
  2424--2432, 2010.

\bibitem{yang2018VisualQuestionAnswer}
Chun-Ju Yang, Kristen Grauman, and Danna Gurari.
\newblock Visual {{Question Answer Diversity}}.
\newblock In {\em {{HCOMP}}}, pages 184--192, 2018.

\end{thebibliography}
